\documentclass{article} %
\usepackage{iclr2027_conference,times}

\usepackage{amsmath,amsfonts,bm}

\def\eqref#1{equation~\ref{#1}}

\def\1{\bm{1}}

\def\mB{{\bm{B}}}
\def\mC{{\bm{C}}}

\DeclareMathAlphabet{\mathsfit}{\encodingdefault}{\sfdefault}{m}{sl}
\SetMathAlphabet{\mathsfit}{bold}{\encodingdefault}{\sfdefault}{bx}{n}

\usepackage{hyperref}
\usepackage{url}
\usepackage{graphicx}
\usepackage{wrapfig}
\usepackage{subcaption}
\usepackage{booktabs}
\usepackage{multirow}
\usepackage{amsmath}
\usepackage{amssymb}
\usepackage{amsfonts}
\usepackage{mathtools}
\usepackage{amsthm}

\usepackage{xcolor}
\usepackage{colortbl}
\usepackage{pifont}
\usepackage{nicefrac}
\usepackage{microtype}

\usepackage{enumitem} %
\usepackage{threeparttable}

\usepackage{float}
\usepackage{bbm}

\usepackage[capitalize,noabbrev]{cleveref}
\usepackage[font=small,labelfont=bf]{caption}

\usepackage[acronym,nomain]{glossaries}
\setacronymstyle{long-short}
\glsdisablehyper %

\definecolor{ourshl}{rgb}{0.90,0.94,1.00}

\hypersetup{
    colorlinks=true,
    citecolor=blue!60!black,
    linkcolor=black,
    urlcolor=teal!60!black,
    pdftitle={RTSM: Adapting Recursive Models to Multivariate Time-Series Classification},
    pdfauthor={Georg Tenzing, Anna Tegon, Luca Benini, Thorir Mar Ingolfsson}
}

\title{RTSM: Adapting Recursive Models to\\Multivariate Time-Series Classification}

\author{%
Georg Tenzing$^{1}$ \And
Anna Tegon$^{1}$ \And
Luca Benini$^{1}$ \And
Thorir Mar Ingolfsson$^{1}$ \And
\makebox[\dimexpr\textwidth-2\tabcolsep][c]{\normalfont $^{1}$Integrated Systems Laboratory, ETH Z{\"u}rich, Switzerland}\\[2pt]
\makebox[\dimexpr\textwidth-2\tabcolsep][c]{\normalfont Corresponding Author: \texttt{thoriri@ethz.ch}}
}
\iclrfinalcopy %
\newacronym{simd}{SIMD}{Single Instruction, Multiple Data}
\newacronym{elu}{ELU}{Exponential Linear Unit}
\newacronym{relu}{ReLU}{Rectified Linear Unit}
\newacronym{gelu}{GELU}{Gaussian Error Linear Unit}
\newacronym{rpr}{RPR}{Random Partition Relaxation}
\newacronym{mac}{MAC}{Multiply Accumulate}
\newacronym{dma}{DMA}{Direct Memory Access}
\newacronym{bmi}{BMI}{Brain--Machine Interface}
\newacronym{bci}{BCI}{Brain--Computer Interface}
\newacronym{smr}{SMR}{Sensory Motor Rythms}
\newacronym{eeg}{EEG}{Electroencephalography}
\newacronym{svm}{SVM}{Support Vector Machine}
\newacronym{svd}{SVD}{Singular Value Decomposition}
\newacronym{evd}{EVD}{Eigendecomposition}
\newacronym{iir}{IIR}{Infinite Impulse Response}
\newacronym{fir}{FIR}{Finite Impulse Response}
\newacronym{fc}{FC}{Fabric Controller}
\newacronym{nn}{NN}{Neural Network}
\newacronym{mrc}{MRC}{Multiscale Riemannian Classifier}
\newacronym{flop}{FLOP}{Floating Point Operation}
\newacronym{sos}{SOS}{Second-Order Section}
\newacronym{ipc}{IPC}{Instructions per Cycle}
\newacronym{tcdm}{TCDM}{Tightly Coupled Data Memory}
\newacronym{fpu}{FPU}{Floating Point Unit}
\newacronym{fma}{FMA}{Fused Multiply Add}
\newacronym{alu}{ALU}{Arithmetic Logic Unit}
\newacronym{dsp}{DSP}{Digital Signal Processing}
\newacronym{gpu}{GPU}{Graphics Processing Unit}
\newacronym{soc}{SoC}{System-on-Chip}
\newacronym{mi}{MI}{Motor-Imagery}
\newacronym{csp}{CSP}{Commmon Spatial Patterns}
\newacronym{fbcsp}{FBCSP}{Filter-Bank \acrlong{csp}}
\newacronym{pulp}{PULP}{parallel ultra-low power}
\newacronym{soa}{SoA}{state-of-the-art}
\newacronym{bn}{BN}{Batch Normalization}
\newacronym{isa}{ISA}{Instruction Set Architecture}
\newacronym{ecg}{ECG}{Electrocardiogram}
\newacronym{ppg}{PPG}{Photoplethysmography}
\newacronym{eog}{EOG}{Electrooculography}
\newacronym{emg}{EMG}{Electromyography}
\newacronym{mcu}{MCU}{microcontroller}
\newacronym{rnn}{RNN}{recurrent neural network}
\newacronym{cnn}{CNN}{convolutional neural network}
\newacronym{tcn}{TCN}{temporal convolutional network}
\newacronym{emu}{EMU}{epilepsy monitoring unit}
\newacronym{pd}{PD}{Parkinson's disease}
\newacronym{ad}{AD}{Alzheimer's disease}
\newacronym{tueg}{TUEG}{Temple University Hospital EEG}
\newacronym{tuh}{TUH}{Temple University Hospital}
\newacronym{tusl}{TUSL}{TUH EEG Slowing Corpus}
\newacronym{tuar}{TUAR}{TUH EEG Artifact Corpus}
\newacronym{tuab}{TUAB}{TUH Abnormal EEG Corpus}
\newacronym{hmc}{HMC}{Haaglanden Medisch Centrum sleep staging database}

\newacronym{fm}{FM}{Foundation Model}
\newacronym{nlp}{NLP}{natural language processing}
\newacronym{gpt}{GPT}{Generative Pre-trained Transformer}
\newacronym{ssm}{SSM}{State Space Model}
\newacronym{moe}{MoE}{Mixture of Experts}
\newacronym{bimamba}{bi-Mamba}{bidirectional Mamba}
\newacronym{lejepa}{LeJEPA}{Latent-Euclidean Joint-Embedding Predictive Architecture}
\newacronym{jepa}{JEPA}{Joint-Embedding Predictive Architecture}
\newacronym{sigreg}{SIGReg}{Sketched Isotropic Gaussian Regularization}
\newacronym{ecf}{ECF}{Empirical Characteristic Function}
\newacronym{cf}{CF}{Characteristic Function}
\newacronym{tsne}{t-SNE}{T-distributed Stochastic Neighbor Embedding}
\newacronym{flops}{FLOPS}{Floating-point Operations Per Second}
\newacronym{oom}{OOM}{Out-Of-Memory}
\newacronym{ssl}{SSL}{Self-Supervised Learning}
\newacronym{fft}{FFT}{Fast Fourier Transform}
\newacronym{ptq}{PTQ}{Post-Training Quantization}
\newacronym{qat}{QAT}{Quantization-Aware Training}
\newacronym{ff}{FF}{Full Fine-Tuning}
\newacronym{fe}{FE}{Frozen Encoder}
\newacronym{lora}{LoRA}{Low-Rank Adaptation}

\newacronym{lda}{LDA}{Linear Discriminant Analysis}
\newacronym{sota}{SOTA}{state-of-the-art}
\newacronym{auroc}{AUROC}{Area Under the Receiver Operating Characteristic}
\newacronym{aupr}{AUPR}{Area Under the Precision-Recall}
\newacronym{balacc}{Bal. Acc}{Balanced Accuracy}

\newacronym{tsc}{TSC}{time-series classification}
\newacronym{rrm}{RRM}{Recursive Reasoning Model}
\newacronym{rtsm}{RTSM}{Recursive Time-Series Model}
\newacronym{trm}{TRM}{Tiny Recursive Model}
\newacronym{hrm}{HRM}{Hierarchical Reasoning Model}
\newacronym{ptrm}{PTRM}{Probabilistic Tiny Recursive Model}
\newacronym{eqr}{EqR}{Equilibrium Reasoners}

\newacronym{swiglu}{SwiGLU}{swish-gated linear unit}
\newacronym{mlp}{MLP}{multi-layer perceptron}

\newacronym{lsst}{LSST}{Large Synoptic Survey Telescope}

\newacronym{rms}{RMS}{root mean square}

\begin{document}

\maketitle
\fancyhead{}
\renewcommand{\headrulewidth}{0pt}

\begin{abstract}
Accurate multivariate time-series classification requires extracting temporal features and integrating information across a sequence. Weight-tied refinement offers a way to repeatedly process these features without allocating new parameters at each stage. Applying this principle to time series requires choosing both a temporal representation and a latent-state organization suited to classification. We introduce the Recursive Time-Series Model (RTSM), which combines a learned convolutional temporal tokenizer, single-state weight-tied refinement, and a pooled classification readout. Across all 30 UEA datasets, RTSM achieves 72.81\% mean accuracy under validation-based configuration and checkpoint selection, the highest observed mean among the 17 evaluated deep-learning models and 1.60 percentage points above MambaSL. Ablations on eight datasets favor the proposed tokenizer over a simple convolutional embedding and the single-state model over the tested two-state variants. The complete classifier reaches 86.10\% mean accuracy versus 84.79\% for a separately trained tokenizer-only classifier. These findings identify temporal encoding and state organization as key design choices for adapting recursive architectures to time series. The resulting classifier combines competitive accuracy with a shared refinement module whose parameter count is independent of computational depth.

\end{abstract}

\section{Introduction}
\label{sec:intro}

Multivariate \gls{tsc} assigns one label to a sequence of observations from multiple variables. Accurate prediction requires extracting temporal patterns and combining evidence across the sequence. The UEA archive illustrates the diversity of this task, with datasets that differ substantially in sequence length, dimensionality, and training-set size~\citep{bagnall2018uea}. Convolutional, attention-based, and state-space models offer different ways to build and process these representations~\citep{mambasl}. Our goal is to achieve strong classification accuracy across this diversity using a refinement module whose parameters are shared across computational depth.

Weight sharing allows repeated computation without allocating a separate set of weights to each processing stage. Recent \glspl{rrm}, including the \gls{hrm}~\citep{wang2025hrm} and \gls{trm}~\citep{jolicoeur2025trm}, use this principle to refine latent states for structured prediction. Adapting these architectures to multivariate \gls{tsc} raises two design questions: how to represent continuous temporal inputs for the shared module, and whether a two-state organization is useful when the output is one sequence-level class label. We therefore examine temporal encoding and latent-state organization together as the basis for an accurate classifier using shared refinement.

\begin{figure*}[t]
    \centering
    \includegraphics[width=\textwidth]{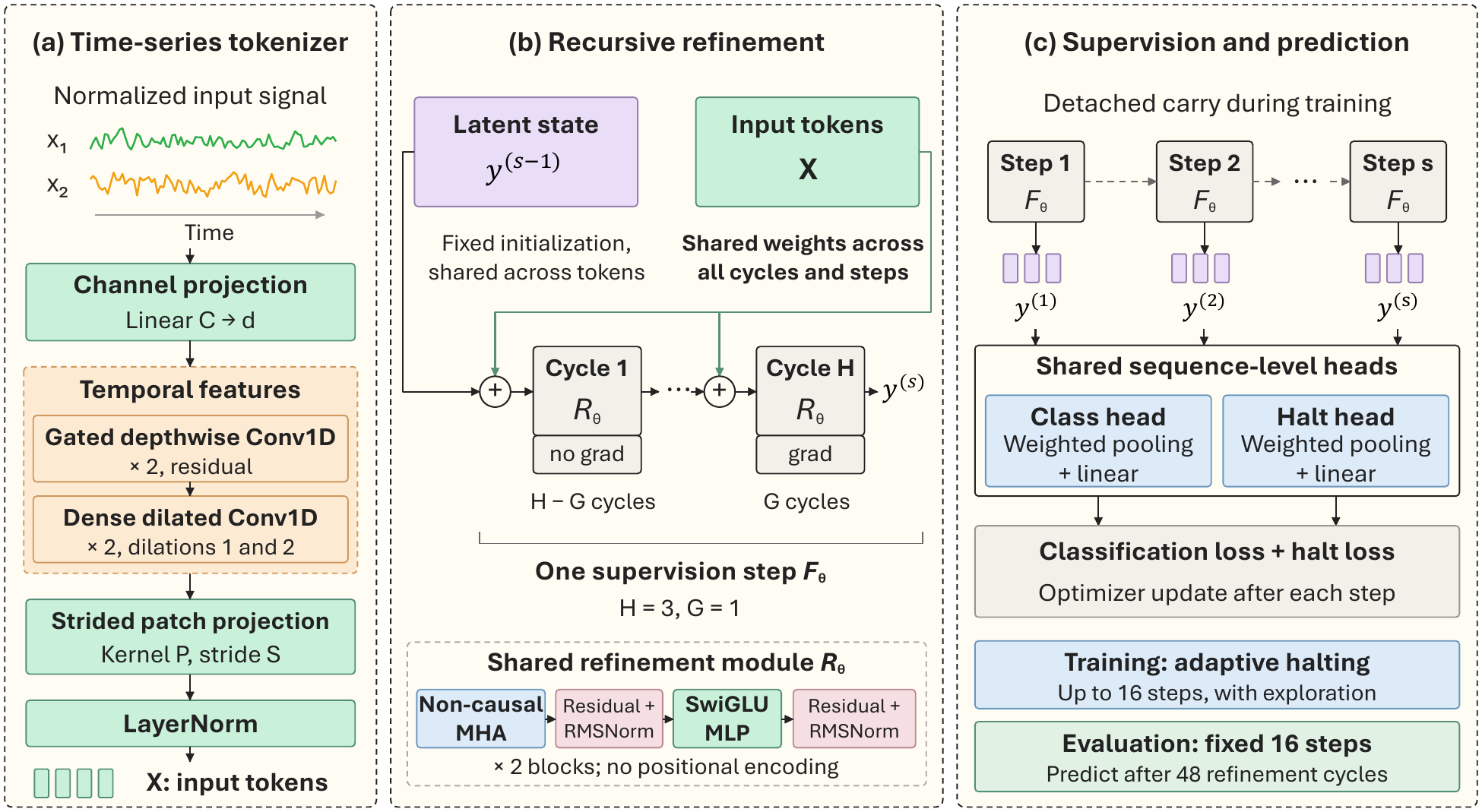} 
    
    \caption{
    \gls{rtsm} encodes a time series into input tokens $X$ (green) and repeatedly updates one latent state $y$ (purple). (a) The tokenizer extracts temporal features before forming patches. Temporal order enters through the tokenizer; refinement uses no positional encoding. (b) Each supervision step maps $y^{(s-1)}$ to $y^{(s)}$ through $H=3$ applications of the shared module $R_\theta$, adding $X$ before every cycle. The initial state broadcasts a fixed random vector $y_{\mathrm{init}}\in\mathbb{R}^d$ across tokens. Training retains gradients through the final $G=1$ cycle.
    (c) Each step receives a classification loss, a halt loss, and an optimizer update; the carried state is detached before the next step. Training uses adaptive halting with exploration, up to 16 steps. Evaluation uses a fixed 16 steps (48 refinement cycles).
    }
    \label{fig:rtsm_architecture}
\end{figure*}

We introduce the \emph{\gls{rtsm}}, which couples a learned temporal encoder to a single-state, weight-tied refinement module. The encoder extracts local temporal features; repeated applications of shared attention combine information across the resulting tokens; and a pooled readout produces the class prediction. Figure~\ref{fig:rtsm_architecture} illustrates the architecture. Our contributions are:

\begin{samepage}
\paragraph{1. Temporal encoding with shared refinement.}
\gls{rtsm} combines a convolutional temporal tokenizer with one evolving latent state and a pooled classification readout. The same attention module is reused throughout refinement. Training builds on \gls{trm}'s deep supervision and adaptive halting, while inference uses a fixed refinement budget (Section~\ref{sec:rtsm}).

\par
\end{samepage}

\paragraph{2. Classification accuracy across the UEA archive.}
Under validation-based configuration and checkpoint selection, \gls{rtsm} achieves $72.81\%$ mean accuracy across all 30 UEA datasets, the highest observed mean among the 17 evaluated deep-learning models and $1.60$ percentage points above MambaSL (Section~\ref{sec:uea_benchmark}).

\paragraph{3. Design insights for recursive time-series classification.}
Across eight datasets, the proposed tokenizer achieves higher mean accuracy than a simple convolutional embedding in each tested comparison. The single-state model also outperforms the tested two-state variants, and the complete classifier improves on a separately trained tokenizer-only model by $1.31$ percentage points. Further ablations examine depth and supervision, followed by an inference-budget analysis on 28 datasets (Sections~\ref{sec:recursive_ablations} and~\ref{sec:inference_scaling}).

\section{Background and related work}
\label{sec:background}

\subsection{Multivariate Time-Series Classification}
Time-series classifiers differ in how they extract temporal features and combine them into a sequence-level prediction. ROCKET and MultiRocket use fixed convolutional filters, pooled feature summaries and linear classifiers~\citep{dempster2020rocket,tan2022multirocket}. Neural approaches learn representations jointly with the prediction objective: ModernTCN uses a convolutional backbone~\citep{luo2024moderntcn}, while MambaSL adapts a selective state-space model for classification~\citep{gu2024mambalineartimesequencemodeling,mambasl}. Attention-based forecasting architectures, including Crossformer, have also been adapted and evaluated for classification within the MambaSL benchmark~\citep{zhang2023crossformer, mambasl}.

Beyond the backbone, the input representation and sequence-level readout are important architectural choices. MambaSL explicitly studies input projection, feature extraction, and aggregation as parts of the classification pipeline~\citep{mambasl}. For a recursive classifier, these choices determine which temporal features condition the latent updates and how the resulting states produce a class prediction. We therefore study the adaptation as a complete classification architecture, assessing temporal tokenization and state structure alongside refinement depth.

\subsection{Recursive Architectures}
\label{sec:bg:rrm_general}
Recursive architectures build computational depth by repeatedly applying shared update modules to an evolving latent representation. Recent \emph{\glspl{rrm}}, including \gls{hrm}~\citep{wang2025hrm} and \gls{trm}~\citep{jolicoeur2025trm}, build on earlier work on adaptive computation and weight-tied depth~\citep{graves2016act, dehghani2019universal}. We describe their design through five components: \emph{latent states}, \emph{recursive update structure}, \emph{gradient schedule}, \emph{deep supervision}, and \emph{halting mechanism}.

\paragraph{Latent states and recursive updates.}
A latent state $y$ holds the representation used for prediction, while optional auxiliary states $\mathbf{z}$ support its updates. An encoder $f_I$ embeds the input $\tilde{x}$, a recursive procedure $F_\theta$ updates the states, and a readout $f_O$ produces the prediction:
\begin{equation}
    x = f_I(\tilde{x}), \qquad
    (y,\mathbf{z}) \leftarrow F_\theta(y,\mathbf{z},x), \qquad
    \hat{y}=f_O(y).
    \label{eq:rrm_update}
\end{equation}
\gls{hrm} and \gls{trm} each use one auxiliary state $z$. In our notation, a call to $F_\theta$ comprises $H_{\mathrm{cycles}}$ cycles, each containing $L_{\mathrm{cycles}}$ updates $z\leftarrow f_L(z,y,x)$ followed by one update $y\leftarrow f_H(y,z)$. \gls{hrm} uses separate networks for the two updates; \gls{trm} shares one network. A single-state formulation omits $\mathbf{z}$, sets $L_{\mathrm{cycles}}=0$, and updates $y$ directly conditioned on $x$.

\paragraph{Gradient schedule.}
Forward computation can extend beyond the portion retained for backpropagation.  We denote by $G_{\mathrm{cycles}}\leq H_{\mathrm{cycles}}$ the number of final cycles tracked for gradients, with earlier cycles evaluated without gradient tracking. \gls{trm} differentiates through the complete final cycle $G_{\mathrm{cycles}}=1$; \gls{hrm} retains only the final $f_L$ and $f_H$
updates. This is analogous to truncating backpropagation through time
~\citep{aicher2020tbptt}, but across refinement cycles. Throughout the paper, $H/L/G$ denotes cycles, inner auxiliary-state updates, and gradient-tracked cycles, respectively.

\paragraph{Deep supervision and halting.}
Intermediate supervision provides a learning signal at successive stages of computation~\citep{lee2015deeply}. In \gls{hrm} and \gls{trm}, each \emph{supervision step} applies the recursive procedure to the states carried from the preceding step and supervises the resulting prediction
\begin{equation}
    \left(y^{(s)},\mathbf{z}^{(s)}\right)
    =
    F_\theta\!\left(y^{(s-1)},\mathbf{z}^{(s-1)},x\right),
    \qquad
    \hat{y}^{(s)}=f_O\!\left(y^{(s)}\right).
    \label{eq:rrm_deep_supervision}
\end{equation}
The carried states are detached from the computation graph between steps. Adaptive halting allows a sample to finish before a maximum budget $S_{\max}$: \gls{hrm} learns halt and continue values through Q-learning, whereas \gls{trm} trains a scalar halt score $q$ to predict whether the current answer is correct. Halting controls the number of supervision steps, separately from the within-step depths $H_{\mathrm{cycles}}$ and $L_{\mathrm{cycles}}$. The stopping rule may differ between training and inference; a fixed inference budget applies the same number of steps to every sample.

\subsubsection{Inference-Time Computation}
\label{sec:bg:rrm_theory}
\label{sec:bg:rrm_inference_scaling}
Inference computation can vary along two axes: \emph{depth} $D$, the number of successive applications of $F_\theta$ along a trajectory, and \emph{breadth} $B$, the number of stochastic trajectories explored. Depth is measured in units corresponding to supervision steps during training. In \gls{eqr}~\citep{huang2026eqr}, models are trained with randomized state initializations and noise injection to support depth and breadth scaling. \gls{ptrm} instead introduces noise only at inference and selects among candidate answers using the learned Q-head~\citep{parviz2026ptrm}. These approaches motivate our characterization of prediction accuracy and latent-state changes under additional deterministic refinement and stochastic exploration (Section~\ref{sec:inference_scaling}).

We next describe how \gls{rtsm} adapts \gls{trm}'s refinement and training framework to sequence classification.
\section{Recursive Time-Series Model}
\label{sec:rtsm}
\gls{rtsm} adapts recursive computation to multivariate classification through temporal encoding, a single latent state, and a sequence-level readout. A convolutional tokenizer first converts the observations into temporal feature vectors, or tokens. A shared attention-based module then updates a latent state containing one vector per token. Each update combines the current state with the tokenizer output, keeping the input features available throughout refinement. A learned pooling operation combines the updated token states into a sequence-level class prediction. Figure~\ref{fig:rtsm_architecture} illustrates these three components.

Building on \gls{trm}~\citep{jolicoeur2025trm}, we supervise predictions at intermediate stages of refinement (deep supervision) and use learned halting during training. At inference, every sample receives a fixed number of refinement steps. Table~\ref{tab:rtsm-adaptation} summarizes the adaptation from \gls{trm}. Section~\ref{sec:rtsm:training} describes the training procedure and evaluation budget, while Appendix~\ref{appendix:rtsm_implementation} provides implementation details and a component-level comparison with \gls{hrm} and \gls{trm}. 

\begin{table}[htbp]
    \centering
    \caption{From \gls{trm} to \gls{rtsm}: the main adaptations and retained training mechanisms. The pooled readout follows MambaSL~\citep{mambasl}; Appendix~\ref{appendix:rtsm_comparison} gives the full component comparison.}
    \label{tab:rtsm-adaptation}
    \small
    \setlength{\tabcolsep}{7.5pt}
    \renewcommand{\arraystretch}{1.15}
    \begin{tabular}{@{}
        >{\raggedright\arraybackslash\bfseries}p{0.16\linewidth}
        >{\raggedright\arraybackslash}p{\dimexpr0.39\linewidth-2\tabcolsep\relax}
        >{\columncolor{teal!5}[\tabcolsep][0pt]\raggedright\arraybackslash}p{\dimexpr0.45\linewidth-2\tabcolsep\relax}@{}}
        \toprule
        Component & \textbf{\gls{trm}} & \textbf{\gls{rtsm}} \\
        \midrule
        Input & Discrete token embeddings & Convolutional temporal tokens from real-valued sequences \\
        \addlinespace[5pt]
        State & Prediction state $y$ and auxiliary state $z$ & One token-level latent state $y$ refined using the input tokens \\
        \addlinespace[5pt]
        Readout & Per-token class logits & Pooled sequence-level class and correctness logits \\
        \addlinespace[5pt]
        \midrule
        Training & Deep supervision; gradients through the final cycle; correctness-based halting & Same recipe, with correctness targets for sequence classification \\
        \bottomrule
    \end{tabular}
\end{table}

\paragraph{Task and notation.}
An input $\tilde{X}\in\mathbb{R}^{C\times T}$ contains $C$ variables observed over $T$ time points, with class label $c\in\{1,\ldots,K\}$. The task is to predict one label for the entire sequence. We use $N$ for the number of tokens and $d$ for their feature width.

\subsection{Time-series Tokenizer}

\label{sec:rtsm:tokenizer}
The tokenizer is a learned temporal encoder that extracts features before attention combines information across tokens. A learned channel projection first combines the observed variables at each time point. Gated residual convolutions then filter features across neighboring time points, using input-dependent gates to modulate the filtered features before adding them back to the representation. Dilated convolutions~\citep{oord2016wavenetgenerativemodelraw} extend the temporal context by filtering at spaced positions. Dense connections~\citep{huang2018denselyconnectedconvolutionalnetworks} give each dilated layer access to the feature maps produced by preceding layers. These operations preserve the sequence length; Appendix~\ref{appendix:rtsm_tokenizer} specifies their computations.

Let $V\in\mathbb{R}^{d\times T}$ denote the resulting feature map. To form tokens, an unpadded convolution maps each patch of $P$ consecutive positions in $V$ to a $d$-dimensional vector, advancing by $S$ positions between patches. Layer normalization gives
\begin{equation}
    X=f_I(\tilde{X})
    =\operatorname{LN}\!\left(\operatorname{Conv}_{P,S}(V)^\top\right)
    \in\mathbb{R}^{N\times d},
    \qquad
    N=1+\left\lfloor\frac{T-P}{S}\right\rfloor .
    \label{eq:rtsm_patch_tokenizer}
\end{equation}
Each token therefore combines features that already incorporate temporal context from the preceding convolutions. Patch length $P$ and stride $S$ determine the token resolution at which the shared refinement module operates and are selected per dataset under the evaluation protocols of Section~\ref{sec:exp}.

\subsection{Single-state Refinement}
\label{sec:rtsm:refinement}
\gls{rtsm} maintains a latent state $y\in\mathbb{R}^{N\times d}$, whose rows represent the input tokens. Refinement repeatedly processes the encoded sequence: its cycles do not advance through the observed time points. The readout below converts the updated state into a class prediction.

One \emph{refinement cycle} adds the input tokens $X$ to the current state and applies the shared attention-based module $R_\theta$. A \emph{step} groups $H_{\mathrm{cycles}}$ such cycles, after which the model produces a prediction that is supervised during training. The reference configuration uses three cycles per step. For cycle $h$ within step $s$, 
\begin{equation}
    y^{(s,h+1)}
    =R_\theta\!\left(y^{(s,h)}+X\right),
    \qquad h=0,\ldots,H_{\mathrm{cycles}}-1 .
    \label{eq:rtsm_single_state}
\end{equation}
Each step starts from the preceding step's terminal state, $y^{(s,0)}=y^{(s-1)}$, and returns $y^{(s)}=y^{(s,H_{\mathrm{cycles}})}$. The input representation remains available at every cycle, while attention mixes information across the token states. All cycles use the same update module and act directly on $y$; the correspondence to \gls{trm}'s two-state formulation is given in Appendix~\ref{appendix:rtsm_comparison}.

For a new sample, a fixed, non-trainable random vector $y_{\mathrm{init}}\in\mathbb{R}^{d}$ is broadcast across the $N$ tokens. The reference module has two blocks, each comprising non-causal multi-head self-attention and a \gls{swiglu} \gls{mlp}, with residual connections and RMSNorm after each sublayer. It uses $d=512$, eight attention heads, and an \gls{mlp} intermediate width of $1536$, with no positional encoding in the refinement module.

\subsection{Sequence-level Readout}
\label{sec:rtsm:readout}

After each refinement step, learned weighted pooling combines the token states into sequence-level outputs. The classification head produces class logits $\ell$, while a second head produces a halt logit $q$, trained to estimate whether the current prediction is correct. This score controls training duration; benchmark inference uses a fixed budget.

Following the pooling construction of MambaSL~\citep{mambasl}, each head scores a token by taking the largest component of a learned projection, then normalizes the scores across tokens. Separate pooling parameters allow classification and halting to emphasize different tokens. For head $r\in\{\mathrm{cls},\mathrm{halt}\}$ and token states $y_1,\ldots,y_N$, the weights and outputs are 
\begin{equation}
    \begin{aligned}
    a_n^r &=
    \operatorname{softmax}_{n}\!
    \left(\max_j[W_r^{\mathrm{pool}}y_n+b_r^{\mathrm{pool}}]_j\right),\\
    \ell &=\sum_{n=1}^{N}a_n^{\mathrm{cls}}W_{\mathrm{cls}}y_n,
    \qquad
    q=w_q^\top\sum_{n=1}^{N}a_n^{\mathrm{halt}}y_n+b_q .
    \end{aligned}
    \label{eq:rtsm_heads}
\end{equation}
Here $j$ indexes the learned pooling channels, and $\operatorname{softmax}_n$ normalizes over the $N$ tokens. The predicted class is $\hat{c}=\arg\max_k\ell_k$. 

With broadcast state initialization, position-independent refinement, and pooled prediction, the classifier is invariant to permutations of the tokenizer outputs. Temporal ordering enters through the tokenizer's convolutional receptive fields.

\subsection{Training and Inference}
\label{sec:rtsm:training}

A training trajectory starts from the broadcast initial state and continues through supervision steps until halting or the maximum budget. In the reference configuration, each step performs two refinement cycles without gradient tracking, followed by one cycle retained for backpropagation ($G_{\mathrm{cycles}}=1$). The resulting state supplies both heads, and the loss at step $s$ supervises the class prediction and its estimated correctness:

\begin{equation}
    \mathcal{L}^{(s)}
    =
    \mathcal{L}_{\mathrm{SM}}\!\left(\ell^{(s)},c\right)
    +
    \lambda_q\,
    \mathcal{L}_{\mathrm{BCE}}\!\left(
        q^{(s)},
        \mathbbm{1}\!\left[\arg\max_k\ell_k^{(s)}=c\right]
    \right).
    \label{eq:rtsm_loss}
\end{equation}
Here $\mathcal{L}_{\mathrm{SM}}$ is the Stablemax cross-entropy~\citep{prieto2025stablemax}, adopted from \gls{trm}~\citep{jolicoeur2025trm}, and $\lambda_q=0.5$ weights the binary correctness objective. An optimizer update follows each supervision step. If the trajectory continues, the terminal latent state is detached from the computation graph and carried forward to initialize the next step. Gradients therefore pass through the final cycle of the current step, but not through earlier cycles or preceding steps.

During training, a positive halt logit can end a sample's trajectory before the maximum $S_{\max}=16$ supervision steps. An exploration rule sometimes delays early halting, as specified in Appendix~\ref{appendix:rtsm_training}. At inference, parameters remain fixed and learned halting is disabled. Every sample receives $S_{\mathrm{eval}}=16$ steps of three cycles each, for 48 applications of $R_\theta$, and the final step supplies the class prediction. In the inference analysis, depth $D$ counts these three-cycle steps, so the benchmark uses $D=16$ (Section~\ref{sec:bg:rrm_inference_scaling}). Evaluation uses an exponential moving average of the trainable parameters with decay $0.999$. Appendix~\ref{appendix:rtsm_training} gives the remaining training settings.

\paragraph{Batch scheduling.}
\label{sec:bg:slot-buffer}
Persistent batch slots accommodate sample-dependent halting. Each slot retains its sample and latent state until the trajectory ends, then accepts another sample. Training progress is counted in completed trajectories per sample, separately from the number of supervised updates. Appendix~\ref{appendix:slot_buffer} specifies how these counts determine the finite training budget.

\paragraph{Computational budget.}
Weight sharing separates model size from refinement depth: increasing the evaluation depth reuses the same parameters. At depth $D$, the reference configuration performs $3D$ applications of its two-block refinement module, corresponding to 96 attention--\gls{mlp} block applications at the benchmark depth $D=16$. The tokenizer-only baseline omits these refinement operations. Computation also depends on the number of tokens $N$, set by the patch length and stride. During training, the truncated gradient schedule retains only the final cycle for backpropagation within each step; the preceding cycles still incur forward computation.

\section{Experiments}
\label{sec:exp}

Our evaluation addresses three questions: how the complete model compares with established classifiers, which configurations support its adaptation to time series, and how its predictions change with additional computation.

\subsection{Classification on the UEA Archive}
\label{sec:uea_benchmark}
\paragraph{Evaluation setup.} %
We evaluate all 30 multivariate \gls{tsc} datasets in the UEA archive~\citep{bagnall2018uea}, training a separate model for each dataset. Our primary evaluation uses an 80/20 stratified split of the official training partition. Model configurations and checkpoints are selected using validation accuracy, and the official test set is reserved for final evaluation. Normalization fitting scopes differ across historical runs and model pipelines, as detailed in Appendix~\ref{appendix:experimental_setup}.

We compare \gls{rtsm} with 16 deep-learning baselines from the MambaSL benchmark~\citep{mambasl}, re-evaluated under this validation protocol using their respective model search spaces. The comparison covers convolutional, attention-based, \gls{mlp}-based, shape-based, and state-space models. Benchmark runs use seed 2021; the primary metric is classification accuracy averaged equally across the 30 datasets. The comparison evaluates complete model-and-training configurations, with their respective optimizers, training settings, and search spaces documented in Appendix~\ref{appendix:experimental_setup}.

\begin{figure}[htbp]
    \centering
        \includegraphics[width=\linewidth]{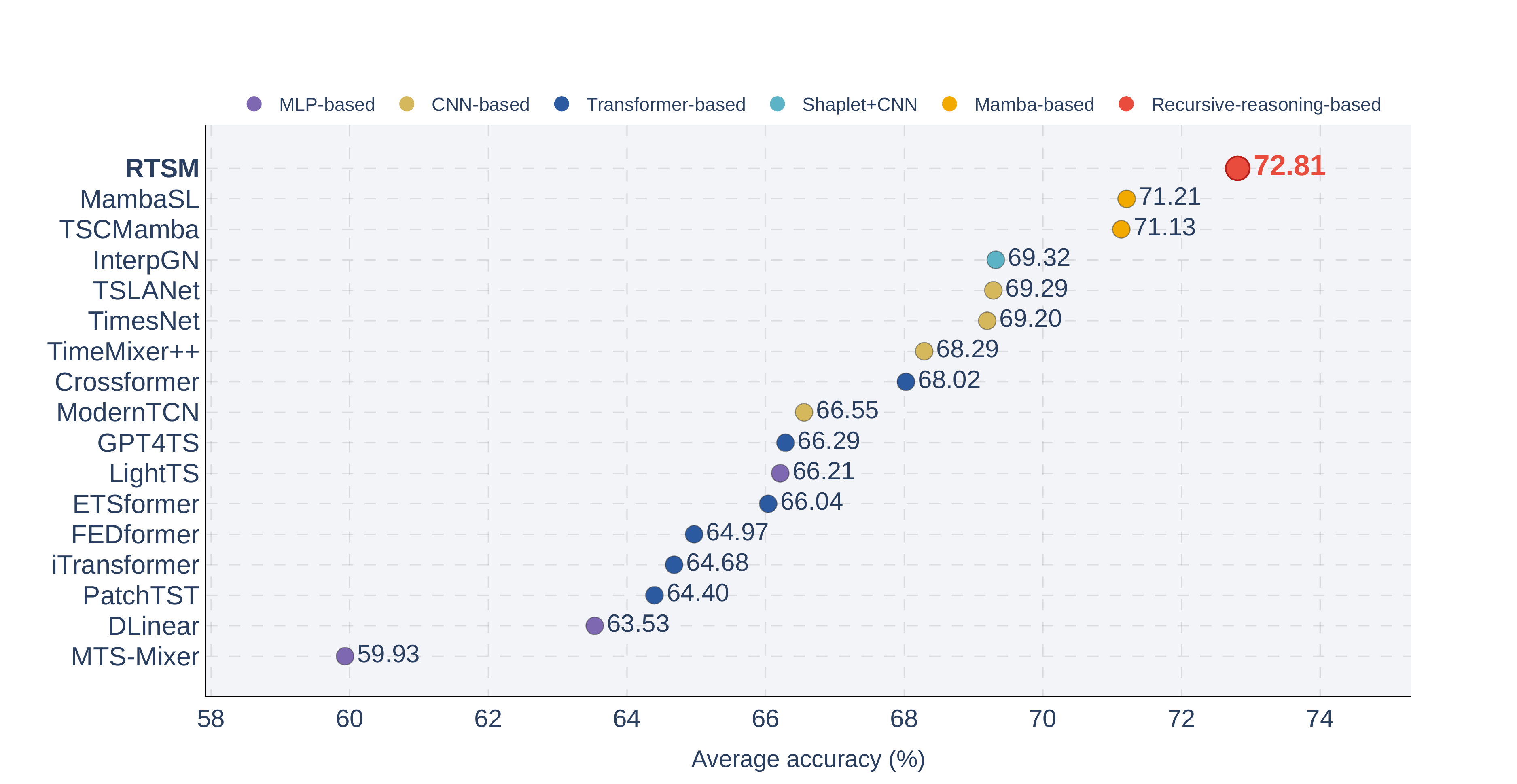}
        \caption{Mean classification accuracy (\%) across all 30 UEA datasets for the 17 evaluated deep-learning models under validation-based configuration and checkpoint selection. Each dataset receives equal weight. Full per-dataset results and pairwise Wilcoxon tests appear in Appendix~\ref{appendix:experimental_results}.}
    \label{fig:validation-avg-acc}
\end{figure}

\paragraph{Results with validation-based selection.} 
\gls{rtsm} achieves the highest observed average accuracy among the 17 evaluated models, reaching $72.81\%$ compared with $71.21\%$ for MambaSL and $71.13\%$ for TSCMamba (Figure~\ref{fig:validation-avg-acc}). Its advantage over MambaSL is $1.60$ percentage points, and it ranks among the top five on 21 of the 30 datasets. The aggregate performance is accompanied by strong results on datasets with different input structures. On the \gls{lsst} dataset, \gls{rtsm} reaches $65.37\%$, exceeding the next-best evaluated model by $7.83$ points. It also achieves the highest accuracy among the evaluated models on EigenWorms, whose sequences contain $17{,}984$ time steps, and PEMS-SF, which has $963$ input channels. These results demonstrate the applicability of the complete architecture across short, long and high-dimensional time series.     

\paragraph{Comparison with the published MambaSL benchmark.}
For continuity with prior results, we also evaluate \gls{rtsm} under the MambaSL protocol~\citep{mambasl}, which trains on the full official training partition and uses the official test set for configuration and checkpoint selection, including early stopping. We reuse the published results for MambaSL and its 20 baselines. \gls{rtsm} reaches $80.82\%$ average accuracy, the highest observed mean in this comparison, exceeding MambaSL's $79.82\%$ by $1.00$ percentage point. Figure~\ref{fig:mambasl-avg-acc} and Table~\ref{tab:full-uea30-result} in Appendix~\ref{appendix:experimental_results} provide the legacy comparison. The protocols are reported separately because their training partitions and selection procedures differ; the validation protocol provides the primary evaluation with test-independent model selection.

\subsection{Architectural and Training Ablations}
\label{sec:recursive_ablations}
The ablations examine the design choices involved in adapting the recursive framework to time-series classification: temporal encoding, state structure, and the refinement schedule. Ablations use eight UEA datasets (CT, DDG, HB, JV, \gls{lsst}, NATO, PD, and SAD), retaining the per-dataset hyperparameters selected under the validation protocol and averaging over seeds 0, 1, and 2. Tables~\ref{tab:state-structure-ablation} and~\ref{tab:recursive-training-ablation} summarize the main comparisons; per-seed results and further variants appear in Appendix~\ref{appendix:trm-analysis-result}.

\paragraph{Temporal encoding and a simpler latent state.}
The proposed tokenizer improves mean accuracy wherever both tokenizers are evaluated (Table~\ref{tab:state-structure-ablation}). With the \gls{rtsm} tokenizer, the single-state model reaches $86.10\%$, compared with $84.98\%$ and $84.29\%$ for the two-state variants. These variants retain the \gls{trm} training recipe; separate update networks capture one architectural feature of \gls{hrm}, rather than reproducing \gls{hrm}'s full procedure. These comparisons favor the proposed temporal encoder paired with a single recurrent state. They compare complete configurations with different parameter counts and computational costs.

\begin{table}[htbp]
    \centering
    \caption{Tokenizer and state structure: mean accuracy (\%) over eight datasets and three seeds. ``None'' applies the prediction head directly to tokenizer outputs. The two-state variants use shared or separate update networks within the \gls{trm} training recipe (Appendix~\ref{appendix:trm-analysis-result}); a dash marks an unevaluated combination.}
    \label{tab:state-structure-ablation}
    \begin{tabular}{lrrrr}
        \toprule
        & \multicolumn{4}{c}{State structure} \\
        \cmidrule(lr){2-5}
        Tokenizer & None & \shortstack{Two-state\\shared $f$} & \shortstack{Two-state\\separate $f_H/f_L$} & Single state \\
        \midrule
        Conv1D & 73.72 & 78.96 & 79.69 & -- \\
        \gls{rtsm}   & 84.79 & 84.98 & 84.29 & \textbf{86.10} \\
        \bottomrule
    \end{tabular}
\end{table}

\begin{figure}[htbp]
    \centering
    \includegraphics[width=\linewidth]{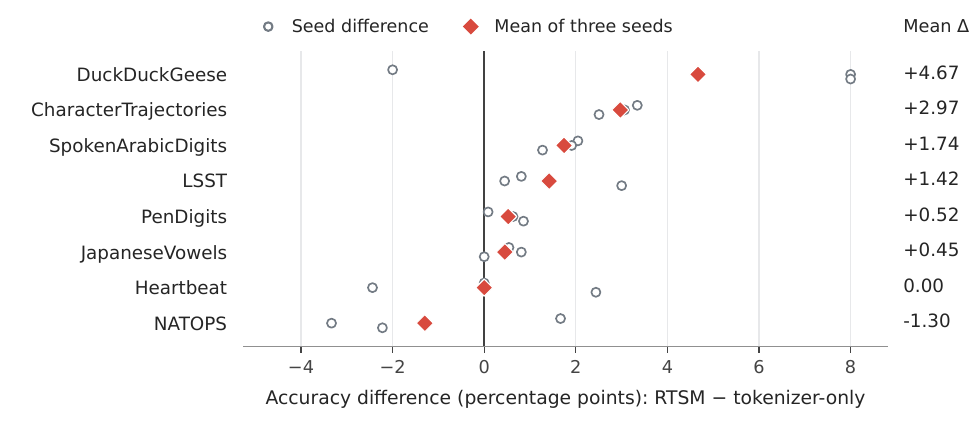}
    \caption{\gls{rtsm} versus the tokenizer-only configuration on all eight ablation datasets. Positive differences favor \gls{rtsm}. Open circles show the accuracy difference for each seed (0, 1, or 2); diamonds and numerical labels show their mean. Small vertical offsets separate the seed markers. Datasets are ordered by their mean difference. Full per-seed accuracies appear in Appendix~\ref{appendix:trm-analysis-result}.}
    \label{fig:tokenizer-dataset-differences}
\end{figure}

\begin{samepage}
\paragraph{Complete model versus tokenizer-only classifier.}
The tokenizer-only classifier trains the temporal encoder and prediction head directly. Using the same tokenizer architecture, the complete single-state model reaches $86.10\pm0.31\%$ versus $84.79\pm0.41\%$ for this baseline, a gain of $1.31$ percentage points. Each seed's score averages the eight datasets equally; $\pm$ denotes the sample standard deviation across the three seed-level scores. Figure~\ref{fig:tokenizer-dataset-differences} shows higher mean accuracy on six datasets, the same mean to two decimal places on Heartbeat, and lower mean accuracy on NATOPS. The overall difference is positive for each of the three seeds. These results favor the complete classifier over direct classification of tokenizer outputs on this subset. Because the comparison changes both attention computation and training, we examine depth and state reset separately below.\par
\end{samepage}

\paragraph{Sequence readout.} The readout ablation yields similar observed mean accuracies: $86.14\%$ with mean pooling and $86.10\%$ with the reference learned weighted pooling (Appendix~\ref{appendix:trm-analysis-result}). \par

\paragraph{Recursive depth and supervision.}
The single-state model maintains similar accuracy across the tested within-step depths: $H=1$--$5$ yields $85.60$--$86.10\%$, with $H=3$ giving the highest mean. At $H=3$, adaptive training with up to 16 supervision steps reaches $86.10\%$, compared with $85.98\%$ for one supervision step and $85.05\%$ when all 16 steps are forced. At $H=1$, updates still recur across supervision steps; at $H=3$, one supervision step contains three refinement cycles. The sweep identifies several effective refinement budgets, with no monotonic accuracy gain from increasing depth. The state-reset variants reinitialize the latent state between supervision steps while retaining refinement within each step. These variants reach $86.00\%$ at $H=1$ and $86.09\%$ at $H=3$, both close to the reference mean of $86.10\%$. Appendix~\ref{appendix:trm-analysis-result} reports all variants.

\begin{table}[htbp]
    \centering
    \caption{Recursive depth and training supervision: mean accuracy (\%) over the same eight datasets and three seeds. Left: vary the number of recursive cycles $H$ within each supervision step, using adaptive supervision capped at 16 steps. Right: vary the supervision schedule at $H=3$.}
    \label{tab:recursive-training-ablation}
    \small
    \setlength{\tabcolsep}{5pt}
    \begin{tabular}{rrrrr@{\hspace{16pt}}rrr}
        \toprule
        \multicolumn{5}{c}{Within-step depth $H$}
        & \multicolumn{3}{c}{Training supervision} \\
        \cmidrule(lr){1-5}\cmidrule(lr){6-8}
        1 & 2 & 3 & 4 & 5 & One step & Adaptive, up to 16 & Fixed 16 \\
        \midrule
        85.77 & 85.60 & \textbf{86.10} & 85.61 & 85.78
        & 85.98 & \textbf{86.10} & 85.05 \\
        \bottomrule
    \end{tabular}
\end{table}

\begin{samepage}
\subsection{Prediction Behavior across Inference Budgets}
\label{sec:inference_scaling}

The preceding ablations change the architecture or training schedule. We now vary the inference budget for fixed trained checkpoints to examine how predictions change with additional computation. This probe covers 28 UEA datasets, excluding EigenWorms and InsectWingbeat because their runs did not complete within the allocated compute budget. We evaluate deterministic trajectories up to depth $D=64$ (three recursive cycles per step) and stochastic trajectories with randomized initialization and between-step noise, using breadths up to $B=512$. Perturbations are introduced only at inference. The comparisons hold checkpoints and preprocessing fixed within the probe; normalization differs from some historical benchmark runs. Appendix~\ref{app:inference_scaling} provides the full protocol, depth--breadth grids, and dataset-level measurements.\par
\end{samepage}

\paragraph{Deterministic trajectories.}
Average accuracy changes from $72.42\%$ at step $1$ to $72.61\%$ at step 16 and $72.84\%$ at step 64. Over the same trajectory, the mean \gls{rms} state-update magnitude decreases from $1.4172$ to $0.0067$. The latent updates therefore become much smaller while aggregate accuracy changes only modestly; extending the trajectory from 16 to 64 steps adds $0.23$ percentage points. Thus, smaller state updates and higher classification accuracy are distinct outcomes of refinement; these measurements characterize state stabilization without establishing convergence to a fixed point.

\paragraph{Stochastic trajectories and selection.}
At depth $D=1$, increasing breadth from $1$ to $512$ raises oracle pass@$B$ accuracy from $72.03\%$ to $77.71\%$. This oracle metric counts a sample as correct if any trajectory predicts its label and therefore measures coverage, not the accuracy of an implementable selector. At $D=64$, the same breadth increase changes oracle accuracy from $72.78\%$ to $73.36\%$. At the maximum budget $(D=64,B=512)$, selecting by the halting score, majority vote, or minimum residual (the smallest recent state changes) yields $72.66\%$, $72.76\%$, and $72.74\%$, respectively, compared with $72.84\%$ for the deterministic trajectory. Within this probe, additional depth yields only modest accuracy changes, and the tested stochastic selectors do not improve on deterministic inference at the maximum budget.

\paragraph{Weight reuse and deployment.}
At its reference inference budget, \gls{rtsm} applies the same 6.82-million-parameter refinement module 48 times. Allocating independent weights to each application would require 327.16 million refinement parameters---a 48-fold difference in refinement-weight storage. At 16 bits per weight, these footprints correspond to 13~MiB and 624~MiB, respectively, excluding the tokenizer, readout, and activations. Reusing the smaller weight set creates an opportunity to reduce off-chip memory traffic when weights remain resident on-chip. This mechanism has demonstrated substantial gains in other recurrent architectures: \citet{diamos2016persistent} report approximately $30\times$ higher recurrent-kernel throughput than a matrix-multiplication implementation on a Titan X \gls{gpu}, using $1{,}152$ hidden units and a batch size of four. These results motivate hardware-aware implementations of \gls{rtsm} for edge devices and \gls{gpu} inference, with benefits depending on memory capacity, activation traffic, and execution scheduling.

\section{Conclusion}
\label{sec:conclusion}

We introduced \gls{rtsm}, an adaptation of recursive architectures to multivariate \gls{tsc} through a learned temporal tokenizer, single-state weight-tied refinement, and a pooled sequence-level readout. Across the 30 UEA datasets, \gls{rtsm} achieves $72.81\%$ average accuracy under validation-based selection, the highest observed mean among the evaluated deep-learning models. Eight-dataset ablations favor the proposed temporal tokenizer over a simple convolutional embedding and the single-state formulation over the tested two-state variants. The complete architecture also improves on a separately trained tokenizer-only classifier by $1.31$ percentage points on this subset.

Depth and state-reset ablations identify several refinement schedules with similar mean accuracy. \gls{rtsm} combines strong classification accuracy with a learned temporal encoder, one latent state, and a refinement module whose parameters are reused across computational depth.

\newpage
\subsubsection*{Reproducibility statement}
The experimental environment, dataset preprocessing, evaluation protocols, hyperparameter configurations and search spaces, random seeds, and checkpoint-selection procedures are documented in Appendix~\ref{appendix:experimental_setup}. All datasets used in this work are publicly available and referenced therein. For evaluations of non-RTSM models, we adapt the implementations and benchmarking infrastructure released with MambaSL~\citep{mambasl}, modifying the evaluation procedure where necessary to follow the validation-based protocol described in this work.

\subsection*{AI use statement}
In this work, we used generative AI tools, including Anthropic Claude and OpenAI Codex, to assist with performance engineering of the experimental code, including refactoring and improving execution efficiency, and to improve the grammar, clarity, and concision of author-written manuscript text. We also used these tools to search for and summarize related work, which helped us identify papers to read. The tools were not used to formulate research questions or hypotheses, design experiments, generate data or experimental results, or draw scientific conclusions. The authors reviewed all retained AI-assisted code changes and validated the final implementation by executing the relevant experiments and inspecting their recorded outputs. All AI-assisted language edits were reviewed and approved by the authors. The authors take responsibility for the final text, code, claims, and artifacts.

\subsubsection*{Acknowledgments}
This work was supported by the Hasler Foundation under the Small Project “Quantized Recursive Inference for Resource-Constrained Edge AI” (project 2026-03-25-875), and by a grant from the Swiss National Supercomputing Centre (CSCS) under project ID lp160 on Alps.

\bibliography{REFERENCES}
\bibliographystyle{iclr2027_conference}

\appendix
\newpage
\section{Experimental setup}
\label{appendix:experimental_setup}

\subsection{Protocol and Implementation Specifications}
\paragraph{Implementation and Computational Environment}
The published MambaSL benchmark results reused in this work were obtained using Python 3.12.8 and PyTorch 2.5.1, primarily on NVIDIA GTX 1080 Ti GPUs, with A100 GPUs used for memory-intensive runs. Classical baselines were evaluated with \texttt{aeon} toolkit~\citep{middlehurst2024aeon}, while most deep-learning models used TSLib or adapted official implementations~\citep{mambasl}. 
\gls{rtsm} was evaluated using a separate training pipeline with persistent batch slots and per-sample trajectory accounting (Section~\ref{sec:bg:slot-buffer}). Samples may halt after different numbers of supervision steps, so the loop tracks completed trajectories separately from optimizer updates. All newly conducted experiments were run on NVIDIA GH200 GPUs~\citep{fusco2024understandingdatamovementtightly}, including \gls{rtsm} under both protocols and the validation-protocol re-evaluation of the deep-learning baselines using the corresponding MambaSL implementations adapted to the validation-based training and selection procedure.

\paragraph{Epoch-Exact Slot-Buffer Accounting}
\label{appendix:slot_buffer}
Adaptive halting allows different samples to complete their training trajectories after different numbers of supervision steps. The slot-buffer scheduler maintains a completion counter $e_i$ for each training sample $i$, with $E$ denoting the target number of epochs. A sample remains in its physical batch slot until it halts; after the corresponding optimizer update, its completed trajectory increments $e_i$ and the slot is refilled from the set of unfinished samples. Samples reaching $e_i=E$ are removed from the refill pool, and training terminates once $e_i=E$ for all samples. If training reaches this budget without protocol-specific early stopping, every sample completes exactly $E$ adaptive trajectories regardless of its halting depth, avoiding depth-dependent over- or undersampling. Non-halted samples retain their slot assignment and latent state across supervision steps, while newly assigned samples are initialized with the model's default latent state.

\paragraph{MambaSL Protocol}
The \emph{MambaSL protocol} follows the evaluation procedure of MambaSL~\citep{mambasl}. Models are trained on the complete official UEA training partition, while the official test partition is evaluated during training and used for model and checkpoint selection as well as early stopping. We reuse the published results for MambaSL and its 20 baselines and evaluate \gls{rtsm} under the same selection procedure. 

\paragraph{Validation Protocol}
The \emph{validation protocol} removes test-set involvement in model selection by partitioning the official UEA training set into an 80/20 stratified train-validation split. The official test partition is reserved for final evaluation, while model configurations and checkpoints are selected by highest validation accuracy, using validation loss as a tie-breaker between configurations. No early stopping is used under this protocol. We re-evaluate the deep-learning baselines and \gls{rtsm} under this protocol while retaining their respective model search spaces. The validation-protocol comparison contains only the deep-learning models; the five non-deep-learning baselines are included in the separate MambaSL-protocol comparison.\paragraph{Normalization.}The reported \gls{rtsm} benchmark includes two historical normalization implementations: statistics were fitted either on the full official training partition before the 80/20 split or on the 80\% training subset after splitting. Both apply the fitted statistics to held-out data; neither fits the \gls{rtsm} normalizer on the official test set. The shared UEA loader used for the re-evaluated baselines standardizes each partition independently, so validation and test statistics are not used to normalize training inputs. Model-specific transforms follow the respective baseline implementations. The comparison therefore reflects the complete model, training, and preprocessing pipelines.

\paragraph{Protocol Hyperparameters}
The basic training settings used across the four experimental groups are summarized in Table~\ref{tab:basic-hyperparams}. For the deep-learning baselines, we retain the standardized settings of MambaSL~\citep{mambasl} and modify only protocol-dependent choices where required by the validation setup. \gls{rtsm} uses its own training configuration under both protocols.

\begin{table}[H]
\centering
\caption{Basic training settings under the MambaSL and validation protocols.}
\label{tab:basic-hyperparams}
\begin{tabular}{lcccc}
\hline
& \multicolumn{2}{c}{\textbf{MambaSL Protocol}}
& \multicolumn{2}{c}{\textbf{Validation Protocol}} \\
Setting
& DL & \gls{rtsm}
& DL & \gls{rtsm} \\
\hline
Train epochs  & 100 & 100 & 100 & 100 \\
Seed          & 2021 & 2021 & 2021 & 2021 \\
\hline
Batch size    & 16$^\dagger$ & dataset-specific & 16 & dataset-specific \\
Learning rate & $10^{-3}$ & $10^{-4}$ & $10^{-3}$ & $10^{-4}$ \\
LR scheduler  & none & constant & none & constant \\
Optimizer     & RAdam & AdamW & RAdam & AdamW \\
Patience      & 10 & 10 & -- & -- \\
Dropout       & 0.1 & 0.0 & 0.1 & 0.0 \\
\hline
\end{tabular}

{\footnotesize $^\dagger$ Smaller batch sizes are used where required by memory constraints.}
\end{table}

\subsection{Shared Benchmark Specification}

\paragraph{Compared Methods}
The validation-protocol comparison includes \gls{rtsm} and 16 deep-learning
baselines from MambaSL~\citep{mambasl}: the linear and \gls{mlp}-based models DLinear~\citep{zeng2023dlinear},
LightTS~\citep{tianping2022lightts}, and MTS-Mixers~\citep{li2023mtsmixer};
models using convolutional, spectral, or multiscale processing, including TimesNet~\citep{wu2023timesnet},
ModernTCN~\citep{luo2024moderntcn}, TSLANet~\citep{eldele2024tslanet},
and TimeMixer++~\citep{wang2025timemixerpp};
the attention-based FEDformer~\citep{zhou2022fedformer},
ETSformer~\citep{woo2022etsformer}, Crossformer~\citep{zhang2023crossformer},
PatchTST~\citep{yuqi2023patchtst}, GPT4TS~\citep{zhou2023onefitsall},
and iTransformer~\citep{liu2024itransformer};
the shape-based InterpGN~\citep{wen2025interpgn};
and TSCMamba~\citep{ahamed2025tscmamba} and MambaSL.
The MambaSL-protocol comparison additionally includes five non-deep-learning
methods: nearest neighbor with dependent multivariate DTW ($\mathrm{DTW}_{D}$)~\citep{shokoohiyekta2017multivariatedtw},
ROCKET~\citep{dempster2020rocket}, HIVE-COTE 2.0~\citep[HC2;][]{middlehurst2021hivecotev2},
Hydra~\citep{dempster2023hydra}, and MultiRocket~\citep{tan2022multirocket}
combined with Hydra.

\paragraph{Metrics}
\label{appendix:metrics}
Following the reporting convention of MambaSL~\citep{mambasl}, we use the following evaluation metrics:
\begin{itemize}[leftmargin=15pt, topsep=-2pt, itemsep=0pt]
\item \textbf{avg. acc:} Mean classification accuracy (\%) across all 30 UEA datasets.
\item \textbf{avg. acc (10):} Mean classification accuracy (\%) across the commonly used 10-dataset UEA subset.
\item \textbf{avg. rank:} Mean per-dataset rank across all 30 UEA datasets. For the reported rank summaries, accuracies are rounded to two decimal places before ranking; ties receive the minimum occupied rank (e.g., 1, 1, 3). Ranks are computed among the models evaluated under each protocol.
\item \textbf{\# of top-$N$:} Number of datasets on which a model ranks among the top $N$ methods under the same rounding and tie convention.
\item \textbf{\# of win/draw:} Number of datasets on which \gls{rtsm} achieves higher or equal accuracy than the compared model.
\item \textbf{\# of loss/draw:} Number of datasets on which \gls{rtsm} achieves lower or equal accuracy than the compared model.
\item \textbf{rank test:} One-sided pairwise $p$-value of the Wilcoxon signed-rank test~\citep{wilcoxon1945test}, with the alternative that \gls{rtsm} has higher per-dataset accuracy than the compared model. These tests accompany the descriptive accuracy summaries.
\end{itemize}

\paragraph{Dataset}
\label{appendix:uea30-details}
The UEA archive~\citep{bagnall2018uea} comprises 30 multivariate time-series classification datasets, summarized in Table~\ref{tab:uea30-description}, with a subset of 10 datasets (\texttt{EC}, \texttt{FD}, \texttt{HW}, \texttt{HB}, \texttt{JV}, \texttt{PEMS}, \texttt{SRS1}, \texttt{SRS2}, \texttt{SAD}, and \texttt{UW}) commonly used in recent TSC benchmarking through \texttt{TSLib}.
\begin{table}[htbp!]
\caption{Overview of the 30 UEA datasets.}
\label{tab:uea30-description}
\centering\resizebox{\columnwidth}{!}{
\addtolength{\tabcolsep}{+2pt}
\begin{tabular}{llrrrrr}
\toprule
\multicolumn{1}{c}{Dataset} &
  \multicolumn{1}{c}{Abbr.} &
  \multicolumn{1}{c}{Train Cases} &
  \multicolumn{1}{c}{Test Cases} &
  \multicolumn{1}{c}{Length} &
  \multicolumn{1}{c}{Channels} &
  \multicolumn{1}{c}{Classes} \\
\midrule
EthanolConcentration      & \texttt{ EC }   & 261   & 263   & 1751    & 3    & 4  \\
FaceDetection             & \texttt{ FD }   & 5890  & 3524  & 62      & 144  & 2  \\
Handwriting               & \texttt{ HW }   & 150   & 850   & 152     & 3    & 26 \\
Heartbeat                 & \texttt{ HB }   & 204   & 205   & 405     & 61   & 2  \\
JapaneseVowels            & \texttt{ JV }   & 270   & 370   & 7--29   & 12   & 9  \\
PEMS-SF                   & \texttt{ PEMS } & 267   & 173   & 144     & 963  & 7  \\
SelfRegulationSCP1        & \texttt{ SRS1 } & 268   & 293   & 896     & 6    & 2  \\
SelfRegulationSCP2        & \texttt{ SRS2 } & 200   & 180   & 1152    & 7    & 2  \\
SpokenArabicDigits        & \texttt{ SAD }  & 6599  & 2199  & 4--93   & 13   & 10 \\
UWaveGestureLibrary       & \texttt{ UW }   & 120   & 320   & 315     & 3    & 8  \\
\midrule
ArticularyWordRecognition & \texttt{ AWR }  & 275   & 300   & 144     & 9    & 25 \\
AtrialFibrillation        & \texttt{ AF }   & 15    & 15    & 640     & 2    & 3  \\
BasicMotions              & \texttt{ BM }   & 40    & 40    & 100     & 6    & 4  \\
CharacterTrajectories     & \texttt{ CT }   & 1422  & 1436  & 60--182 & 3    & 20 \\
Cricket                   & \texttt{ CR }   & 108   & 72    & 1197    & 6    & 12 \\
DuckDuckGeese             & \texttt{ DDG }  & 50    & 50    & 270     & 1345 & 5  \\
EigenWorms                & \texttt{ EW }   & 128   & 131   & 17984   & 6    & 5  \\
Epilepsy                  & \texttt{ EP }   & 137   & 138   & 206     & 3    & 4  \\
ERing                     & \texttt{ ER }   & 30    & 270   & 65      & 4    & 6  \\
FingerMovements           & \texttt{ FM }   & 316   & 100   & 50      & 28   & 2  \\
HandMovementDirection     & \texttt{ HMD }  & 160   & 74    & 400     & 10   & 4  \\
InsectWingbeat            & \texttt{ IW }   & 25000 & 25000 & 2--22   & 200  & 10 \\
Libras                    & \texttt{ LIB }  & 180   & 180   & 45      & 2    & 15 \\
LSST                      & \texttt{ LSST } & 2459  & 2466  & 36      & 6    & 14 \\
MotorImagery              & \texttt{ MI }   & 278   & 100   & 3000    & 64   & 2  \\
NATOPS                    & \texttt{ NATO } & 180   & 180   & 51      & 24   & 6  \\
PenDigits                 & \texttt{ PD }   & 7494  & 3498  & 8       & 2    & 10 \\
PhonemeSpectra            & \texttt{ PS }   & 3315  & 3353  & 217     & 11   & 39 \\
RacketSports              & \texttt{ RS }   & 151   & 152   & 30      & 6    & 4  \\
StandWalkJump             & \texttt{ SWJ }  & 12    & 15    & 2500    & 4    & 3  \\ 
\midrule   
Min                       &                 & 12    & 15    & 2       & 2    & 2  \\ 
Max                       &                 & 25000 & 25000 & 17984   & 1345 & 39 \\ 
\bottomrule
\end{tabular}
\addtolength{\tabcolsep}{-2pt}
}
\end{table}

\newpage
\subsection{Model hyperparameters for grid search}
For the baseline models, we retain the hyperparameter search spaces defined in the MambaSL benchmark~\citep{mambasl}, which were derived primarily from the respective original papers, official implementations, and \texttt{TSLib} configurations. For forecasting-origin and foundation models without dedicated classification settings, MambaSL additionally considered more compact configurations suitable for the generally smaller UEA datasets. The resulting searches cover more than 200 candidate configurations per model-dataset pair in most cases. We use the same search spaces under both evaluation protocols and list them below together with the corresponding \gls{rtsm} hyperparameters.

\begin{itemize}[leftmargin=15pt, itemsep=8pt]
    \item \textbf{Non-DL}
    \begin{itemize}[leftmargin=15pt, topsep=2pt, itemsep=5pt]
        \item $\text{DTW}_D$: fixed implementations (no learnable hyperparameters).
        \item  ROCKET, HC2, Hydra, MR+Hydra: 3 combinations
            \begin{itemize}[leftmargin=15pt, topsep=0pt, itemsep=2pt]
                \item \texttt{seed}: 0, 1, 2
                \item 
                Other parameters use the fixed configurations reported in the MambaSL benchmark.
            \end{itemize}
    \end{itemize}

    \item \textbf{Linear and \gls{mlp}-based models}
    \begin{itemize}[leftmargin=15pt, topsep=2pt, itemsep=5pt]
        \item DLinear: 15 combinations
            \begin{itemize}[leftmargin=15pt, topsep=0pt, itemsep=2pt]
                \item \texttt{moving\_avg}: 0.5\%, 1\%, 2\%, 3\%, 4\%, 5\%, 
                10\%, 15\%, 20\%, 25\%, 30\%, 35\%, 40\%, 45\%, 50\% of sequence length
                \item 
                Only the moving-average window is varied, giving 15 configurations.
            \end{itemize}
        
        \item LightTS: 100 combinations
            \begin{itemize}[leftmargin=15pt, topsep=0pt, itemsep=2pt]
                \item \texttt{d\_model}: 32, 64, 128, 256, 512
                \item \texttt{chunk\_size}: 1/21\,--\,1/2 of sequence length
                \item 
                The two searched hyperparameters give 100 configurations.
            \end{itemize}

        \item MTS-Mixers: 256 combinations
            \begin{itemize}[leftmargin=15pt, topsep=0pt, itemsep=2pt]
                \item \texttt{e\_layers}: 2
                \item \texttt{d\_model}: 128, 256, 512, 1024
                \item \texttt{d\_ff}: 0, 2, 4, 8, 16, 32, 64, 128
                \item \texttt{fac\_C}: 0 (False) if \texttt{d\_ff} is 0, else 1 (True)
                \item \texttt{down\_sampling\_window}: 0, 1\%, 2\%, 3\%, 5\%, 7.5\%, 10\%, 12.5\% of sequence length
                \item \texttt{fac\_T}: 0 (False) if \texttt{down\_sampling\_window} is 0, else 1 (True)
                \item \texttt{use\_norm}: 1 (True)
            \end{itemize}
        
    \end{itemize}
    
    \newpage
    
    \item \textbf{Convolutional, spectral, and multiscale models}
    \begin{itemize}[leftmargin=15pt, topsep=2pt, itemsep=5pt]
        \item TimesNet: 252 combinations
            \begin{itemize}[leftmargin=15pt, topsep=0pt, itemsep=2pt]
                \item \texttt{e\_layers}: 2, 3, 4
                \item \texttt{(d\_model, d\_ff)}: (8,16), (8,32), (8,64),
                                                  (16,16), (16,32), (16,64),
                                                  (32,32), (32,64), (32,128),
                                                  (64,64), (64,128), (64,256),
                                                  (128,128), (128,256)
                \item \texttt{top\_k}: 1, 2, 3
                \item \texttt{num\_kernels}: 4, 6
            \end{itemize}
        
        \item ModernTCN: 252 combinations
            \begin{itemize}[leftmargin=15pt, topsep=0pt, itemsep=2pt]
                \item \texttt{ffn\_ratio}: 1, 2, 4
                \item \texttt{patch\_size}: 2.5\%, 5\%, 7.5\%, 10\%, 15\%, 20\%, 25\% of sequence length
                \item \texttt{patch\_stride}: 50\% of \texttt{patch\_size}
                \item \texttt{(num\_blocks, large\_size, small\_size, dims)}:
                (``1'', ``13'', ``5'', ``32''), (``1'', ``13'', ``5'', ``64''), (``1'', ``13'', ``5'', ``128''), (``1'', ``13'', ``5'', ``256''),
                (``1 1'', ``9 9'', ``5 5'', ``32 64''), (``1 1'', ``9 9'', ``5 5'', ``64 128''), (``1 1'', ``9 9'', ``5 5'', ``128 256''),
                (``1 1'', ``13 13'', ``5 5'', ``32 64''), (``1 1'', ``13 13'', ``5 5'', ``64 128''), (``1 1'', ``13 13'', ``5 5'', ``128 256''),
                (``1 1 1'', ``9 9 9'', ``5 5 5'', ``32 64 128''), (``1 1 1'', ``13 13 13'', ``5 5 5'', ``32 64 128'')
            \end{itemize}
        
        \item TSLANet: 252 combinations
            \begin{itemize}[leftmargin=15pt, topsep=0pt, itemsep=2pt]
                \item \texttt{depth}: 1, 2, 3
                \item \texttt{emb\_dim}: 32, 64, 128, 256
                \item \texttt{mlp\_ratio}: 1, 2, 3
                \item \texttt{patch\_size}: 2.5\%, 5\%, 7.5\%, 10\%, 15\%, 20\%, 25\% of sequence length
                \item \texttt{patch\_stride}: 50\% of \texttt{patch\_size}
                \item \texttt{masking\_ratio}: 0.4
                \item \texttt{ICB}: 1 (True)
                \item \texttt{ASB}: 1 (True)
                \item \texttt{adaptive\_filter}: 1 (True)
                \item \texttt{load\_from\_pretrained}: 1 (True)
                \item \texttt{pretrain\_lr}: 0.001
                \item \texttt{pretrain\_epoch}: 50
                
            \end{itemize}
        
        \item TimeMixer++: 216 combinations
            \begin{itemize}[leftmargin=15pt, topsep=0pt, itemsep=2pt]
                \item \texttt{down\_sampling\_method}: ``conv''
                \item \texttt{down\_sampling\_layers}: 1, 2, 3
                \item \texttt{down\_sampling\_window}: 2
                \item \texttt{num\_kernels}: 6
                \item \texttt{e\_layers}: 1, 2, 3, 4
                \item \texttt{d\_model}: 16, 32, 64
                \item \texttt{d\_ff}: 16, 32, 64
                \item \texttt{n\_heads}: 8
                \item \texttt{top\_k}: 2, 3
            \end{itemize}
        
    \end{itemize}
    
    \newpage

    \item \textbf{Transformer-based models}
    \begin{itemize}[leftmargin=15pt, topsep=2pt, itemsep=5pt]
        \item FEDformer: 250 combinations
            \begin{itemize}[leftmargin=15pt, topsep=0pt, itemsep=2pt]
                \item \texttt{moving\_avg}: 0.5\%, 1\%, 2\%, 3\%, 5\%, 10\%, 20\%, 30\%, 40\%, 50\% of sequence length
                \item \texttt{e\_layers}: 2
                \item \texttt{d\_model}: 32, 64, 128, 256, 512
                \item \texttt{d\_ff}: 128, 256, 512, 1024, 2048
                \item \texttt{n\_heads}: 8
            \end{itemize}
        \item ETSformer: 240 combinations
            \begin{itemize}[leftmargin=15pt, topsep=0pt, itemsep=2pt]
                \item \texttt{e\_layers}: 1, 2
                \item \texttt{d\_model}: 32, 64, 128, 256, 512
                \item \texttt{d\_ff}: 64, 128, 256, 512, 1024, 2048
                \item \texttt{n\_heads}: 8
                \item \texttt{top\_k}: 1, 2, 3, 4
            \end{itemize}
        
        \item Crossformer: 288 combinations
            \begin{itemize}[leftmargin=15pt, topsep=0pt, itemsep=2pt]
                \item \texttt{e\_layers}: 1, 2, 3
                \item \texttt{d\_model}: 32, 64, 128, 256
                \item \texttt{d\_ff}: 64, 128, 256, 512
                \item \texttt{n\_heads}: 4
                \item \texttt{factor}: 3, 10
                \item \texttt{seg\_len}: 6, 12, 24 (change the code of \texttt{TSLib} to modify the fixed value)
            \end{itemize}
        
        \item PatchTST: 252 combinations
            \begin{itemize}[leftmargin=15pt, topsep=0pt, itemsep=2pt]
                \item \texttt{e\_layers}: 1, 2, 3
                \item \texttt{d\_model}: 16, 32, 64, 128
                \item \texttt{d\_ff}: 64, 128, 256
                \item \texttt{n\_heads}: 4 if $\texttt{d\_model} \in \{16, 32\}$ else 16
                \item \texttt{patch\_size}: 2.5\%, 5\%, 7.5\%, 10\%, 15\%, 20\%, 25\% of sequence length
                \item \texttt{patch\_stride}: 50\% of \texttt{patch\_size}
            \end{itemize}

        \item GPT4TS: 252 combinations
            \begin{itemize}[leftmargin=15pt, topsep=0pt, itemsep=2pt]
                \item \texttt{e\_layers}: 3, 4, 5, 6
                \item \texttt{d\_model}: 768 (fixed since the model use pretrained GPT-2)
                \item \texttt{d\_ff}: 768 (fixed since the model use pretrained GPT-2)
                \item \texttt{patch\_size}: 1/25\,--\,1/5 of sequence length
                \item \texttt{patch\_stride}: 25\%, 50\%, 100\% of \texttt{patch\_size}
            \end{itemize}
        
        \item iTransformer: 240 combinations
            \begin{itemize}[leftmargin=15pt, topsep=0pt, itemsep=2pt]
                \item \texttt{e\_layers}: 1, 2, 3, 4
                \item \texttt{d\_model}: 64, 128, 256, 512, 1024, 2048
                \item \texttt{d\_ff}: 64, 128, 256, 512, 1024
                \item \texttt{n\_heads}: 8
                \item \texttt{factor}: 1, 3
            \end{itemize}
        
    \end{itemize}
    
    \newpage

    \item \textbf{Shape-based model}
    \begin{itemize}[leftmargin=15pt, topsep=2pt, itemsep=5pt]
        \item InterpGN: 243 combinations
            \begin{itemize}[leftmargin=15pt, topsep=0pt, itemsep=2pt]
                \item \texttt{dnn\_type}: ``FCN''
                \item \texttt{num\_shaplet}: 5, 10, 15
                \item \texttt{lambda\_div}: 0, 0.1, 1.0
                \item \texttt{lambda\_reg}: 0, 0.1, 1.0
                \item \texttt{epsilon}: 0.5, 1.0, 2.0
                \item \texttt{gating\_value}: 0.5, 0.75, 1.0
                \item \texttt{distance\_func}: ``euclidean''
                \item \texttt{memory\_efficient}: False
                \item \texttt{sbm\_cls}: ``linear''
                
            \end{itemize}
        
    \end{itemize}

    \item \textbf{Mamba-based models}
    \begin{itemize}[leftmargin=15pt, topsep=2pt, itemsep=5pt]
        \item TSCMamba
            \begin{itemize}[leftmargin=15pt, topsep=0pt, itemsep=2pt]
                \item \texttt{d\_model}: 64, 128, 256
                \item \texttt{e\_layers}: 1, 2
                \item \texttt{expand}: 1, 2
                \item \texttt{d\_conv}: 2, 4
                \item \texttt{d\_ff} (\texttt{d\_state}): 32, 64, 128
                \item \texttt{no\_rocket}: 0 (False)
                \item \texttt{half\_rocket}: 0 (False), 1 (True)
                \item \texttt{additive\_fusion}: 0 (multiplicative fusion), 1 (additive fusion)
                \item \texttt{max\_pooling}: 0 (avg pooling), 1 (max pooling)
                \item \texttt{channel\_token\_mixing}: 0 (False)
                \item \texttt{only\_forward\_scan}: 0 (False)
                \item \texttt{flip\_dir}: 2 (vertical flip)
                \item \texttt{reverse\_flip}: 0 (False)
                \item \texttt{patch\_size}: 8
                \item \texttt{patch\_stride}: 8
                \item \texttt{rescale\_size}: 64
                \item \texttt{variation}: 64
                \item \texttt{initial\_focus}: 1.0

            \end{itemize}
        
        \item MambaSL: 240 combinations
            \begin{itemize}[leftmargin=15pt, topsep=0pt, itemsep=2pt]
                \item \texttt{d\_model} ($d_{\mathrm{m}}$): 32, 64, 128, 256, 512, 1024
                \item \texttt{d\_ff} ($d_{\mathrm{s}}$): 1, 2, 4, 8, 16
                \item \texttt{tv\_dt} ($\theta_\Delta$): 0 (False), 1 (True)
                \item \texttt{tv\_B} ($\theta_\mB$): 0 (False), 1 (True)
                \item \texttt{tv\_C} ($\theta_\mC$): 0 (False), 1 (True)
                \item \texttt{use\_D}: 0 (False)
                \item \texttt{expand}: 1
                \item \texttt{d\_conv}: 4
            \end{itemize}
    \end{itemize}

    \item \textbf{Recursive-reasoning-based models}
    \begin{itemize}[leftmargin=15pt, topsep=2pt, itemsep=5pt]
        \item \gls{rtsm} (ours): 240 combinations
            \begin{itemize}[leftmargin=15pt, topsep=0pt, itemsep=2pt]
                \item \texttt{Batch size}: 8, 16, 32, 64
                \item \texttt{patch\_length}: 1, 2, 3, 4, 8, 10, 16, 24, 48, 64, 128, 400
                \item \texttt{patch\_stride}: 1, 2, 3, 8, 10
            \end{itemize}
    \end{itemize}    
\end{itemize}

\newpage
\section{Experimental Results}
\label{appendix:experimental_results}

\subsection{Full Results on the UEA benchmark under MambaSL protocol}
Table~\ref{tab:full-uea30-result} reports the complete per-dataset results under the MambaSL protocol. The baseline results are taken from the original MambaSL benchmark~\citep{mambasl}, while \gls{rtsm} is evaluated under the same training and selection procedure. Full protocol and implementation details are provided in Appendix~\ref{appendix:experimental_setup}.
\begin{figure}[htbp]
    \centering
    \includegraphics[width=\linewidth]{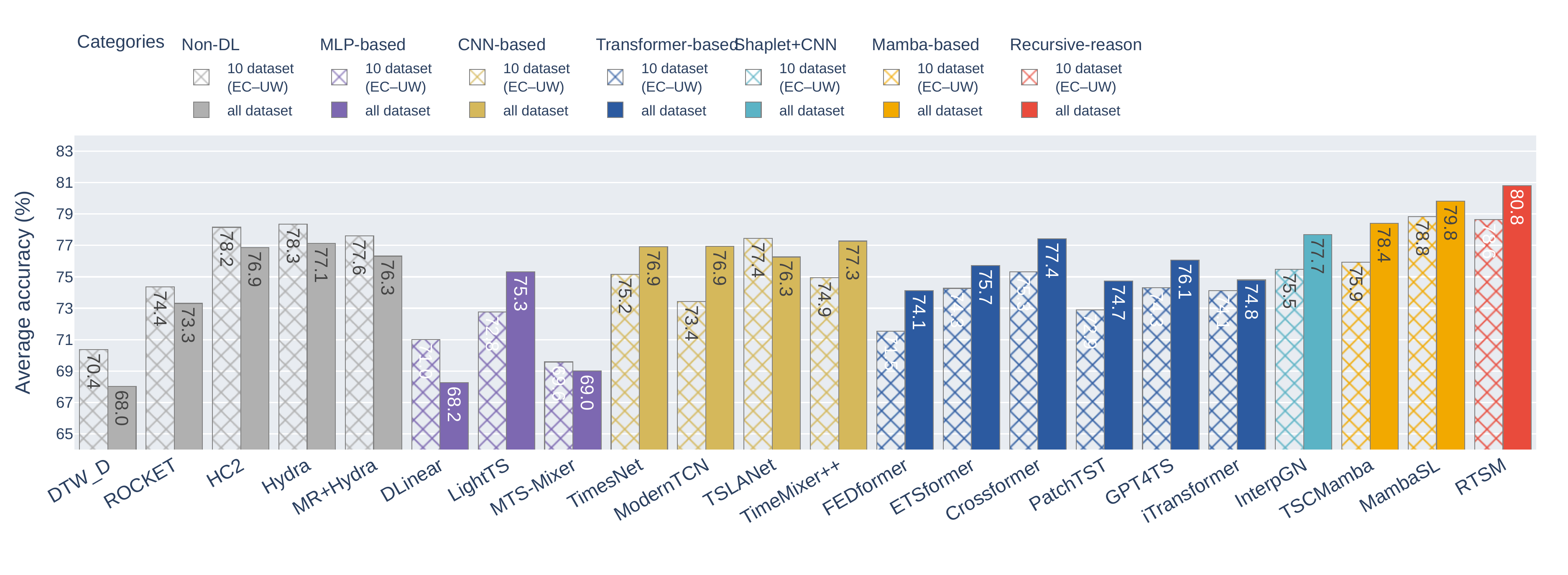}
    \caption{Mean classification accuracy (\%) under the MambaSL protocol, which uses official test-set performance for configuration and checkpoint selection. The comparison includes RTSM, 16 other deep-learning models, and five non-deep-learning baselines. Solid bars show averages across all 30 UEA datasets; hatched bars show the 10-dataset subset. Each dataset receives equal weight within its respective average.}
    \label{fig:mambasl-avg-acc}
\end{figure}

\begin{table}[htbp!]
\centering
\caption{Classification accuracy\,(\%) of models on the 30 datasets of the UEA archive under the MambaSL protocol.
The best and the second-best are highlighted in bold and underline, respectively.}
\label{tab:full-uea30-result}

\begin{subtable}{\columnwidth}
\caption{Classification results from $\text{DTW}_D$ to TimeMixer++.}
\resizebox{\columnwidth}{!}{
\addtolength{\tabcolsep}{-2pt}
\begin{threeparttable}
\begin{tabular}{lrrrrrrrrrrrr}
\toprule
 &
  \multicolumn{5}{c}{non-DL} &
  \multicolumn{3}{c}{Linear/MLP} &
  \multicolumn{4}{c}{Convolutional/spectral/multiscale} \\ \cmidrule(r){2-6} \cmidrule(lr){7-9} \cmidrule(l){10-13}
\multicolumn{1}{c}{} &
  \multicolumn{1}{c}{\small{\begin{tabular}[c]{@{}c@{}}$\text{DTW}_D$\\ \citeyearpar{shokoohiyekta2017multivariatedtw}\end{tabular}}} &
  \multicolumn{1}{c}{\small{\begin{tabular}[c]{@{}c@{}}ROCKET\\ \citeyearpar{dempster2020rocket}\end{tabular}}} &
  \multicolumn{1}{c}{\small{\begin{tabular}[c]{@{}c@{}}HC2\\ \citeyearpar{middlehurst2021hivecotev2}\end{tabular}}} &
  \multicolumn{1}{c}{\small{\begin{tabular}[c]{@{}c@{}}Hydra\\ \citeyearpar{dempster2023hydra}\end{tabular}}} &
  \multicolumn{1}{c}{\small{\begin{tabular}[c]{@{}c@{}}MR+\\Hydra\\ \citeyearpar{dempster2023hydra}\end{tabular}}} &
  \multicolumn{1}{c}{\small{\begin{tabular}[c]{@{}c@{}}DLinear\\ \citeyearpar{zeng2023dlinear}\end{tabular}}} &
  \multicolumn{1}{c}{\small{\begin{tabular}[c]{@{}c@{}}LightTS\\ \citeyearpar{tianping2022lightts}\end{tabular}}} &
  \multicolumn{1}{c}{\small{\begin{tabular}[c]{@{}c@{}}MTS-\\Mixers\\ \citeyearpar{li2023mtsmixer}\end{tabular}}} &
  \multicolumn{1}{c}{\small{\begin{tabular}[c]{@{}c@{}}TimesNet\\ \citeyearpar{wu2023timesnet}\end{tabular}}} &
  \multicolumn{1}{c}{\small{\begin{tabular}[c]{@{}c@{}}Modern\\TCN\\ \citeyearpar{luo2024moderntcn}\end{tabular}}} &
  \multicolumn{1}{c}{\footnotesize{\begin{tabular}[c]{@{}c@{}}TSLANet\\ \citeyearpar{eldele2024tslanet}\end{tabular}}} &
  \multicolumn{1}{c}{\small{\begin{tabular}[c]{@{}c@{}}Time\\Mixer++\\ \citeyearpar{wang2025timemixerpp}\end{tabular}}} \\ \midrule
EC &
  32.319 &
  44.487 &
  52.852 &
  {\underline{54.373}} &
  \textbf{60.076} &
  31.179 &
  32.319 &
  30.798 &
  33.460 &
  32.319 &
  31.939 &
  34.601\; \\
FD &
  52.866 &
  61.606 &
  61.379 &
  60.982 &
  61.294 &
  69.495 &
  68.473 &
  68.530 &
  {\underline{70.148}} &
  66.998 &
  66.969 &
  69.665\; \\
HW &
  60.706 &
  59.294 &
  56.000 &
  56.353 &
  53.529 &
  23.765 &
  25.059 &
  19.647 &
  37.529 &
  30.118 &
  {\underline{62.000}} &
  33.647\; \\
HB &
  71.707 &
  75.610 &
  78.537 &
  76.585 &
  77.073 &
  76.585 &
  77.561 &
  79.024 &
  \textbf{83.902} &
  78.049 &
  79.512 &
  80.488\; \\
JV &
  95.946 &
  83.784 &
  98.378 &
  97.838 &
  98.378 &
  96.486 &
  96.757 &
  95.946 &
  98.649 &
  98.649 &
  98.919 &
  {\underline{99.189}}\; \\
PEMS &
  71.098 &
  84.393 &
  {\underline{97.688}} &
  \textbf{98.266} &
  82.659 &
  82.659 &
  87.861 &
  82.081 &
  87.861 &
  86.705\tnote{A} &
  89.017 &
  89.017\; \\
SRS1 &
  77.474 &
  85.666 &
  89.078 &
  86.689 &
  \textbf{94.881} &
  92.833 &
  92.833 &
  78.840 &
  92.150 &
  93.515 &
  89.761 &
  92.150\; \\
SRS2 &
  53.889 &
  55.556 &
  54.444 &
  58.889 &
  55.000 &
  57.222 &
  59.444 &
  58.333 &
  60.000 &
  60.000 &
  {\underline{63.889}} &
  60.556\; \\
SAD &
  97.226 &
  99.227 &
  99.136 &
  99.136 &
  99.045 &
  96.953 &
  98.226 &
  99.454 &
  99.545 &
  99.500 &
  98.045 &
  99.545\; \\
UW &
  90.313 &
  94.063 &
  94.063 &
  {\underline{94.375}} &
  94.063 &
  82.813 &
  89.063 &
  83.125 &
  88.438 &
  88.438 &
  {\underline{94.375}} &
  90.625\; \\
AWR &
  98.667 &
  \textbf{99.333} &
  \textbf{99.333} &
  \textbf{99.333} &
  \textbf{99.333} &
  97.667 &
  98.333 &
  {\underline{99.000}} &
  98.667 &
  98.667 &
  \textbf{99.333} &
  {\underline{99.000}}\; \\
AF &
  20.000 &
  6.667 &
  20.000 &
  26.667 &
  6.667 &
  53.333 &
  {\underline{73.333}} &
  60.000 &
  66.667 &
  {\underline{73.333}} &
  66.667 &
  66.667\; \\
BM &
  {\underline{97.500}} &
  \textbf{100.000} &
  \textbf{100.000} &
  \textbf{100.000} &
  \textbf{100.000} &
  87.500 &
  \textbf{100.000} &
  \textbf{100.000} &
  \textbf{100.000} &
  \textbf{100.000} &
  \textbf{100.000} &
  \textbf{100.000}\; \\
CT &
  98.816 &
  99.304 &
  99.304 &
  99.164 &
  99.373 &
  98.120 &
  98.955 &
  97.981 &
  98.747 &
  99.304 &
  99.164 &
  99.373\; \\
CR &
  \textbf{100.000} &
  \textbf{100.000} &
  \textbf{100.000} &
  \textbf{100.000} &
  {\underline{98.611}} &
  91.667 &
  93.056 &
  97.222 &
  97.222 &
  97.222 &
  91.667 &
  {\underline{98.611}}\; \\
DDG &
  58.000 &
  52.000 &
  70.000 &
  68.000 &
  60.000 &
  66.000 &
  56.000 &
  68.000\tnote{8} &
  68.000 &
  66.000\tnote{A} &
  74.000 &
  74.000\; \\
EW &
  61.832 &
  91.603 &
  {\underline{97.710}} &
  \textbf{98.473} &
  \textbf{98.473} &
  42.748\tnote{A} &
  54.198\tnote{8} &
  66.412 &
  63.359 &
  64.122 &
  81.679 &
  61.069\tnote{A}\; \\
EP &
  96.377 &
  {\underline{99.275}} &
  \textbf{100.000} &
  \textbf{100.000} &
  \textbf{100.000} &
  65.217 &
  97.101 &
  66.667 &
  96.377 &
  96.377 &
  \textbf{100.000} &
  94.928\; \\
ER &
  91.481 &
  {\underline{98.889}} &
  {\underline{98.889}} &
  {\underline{98.889}} &
  \textbf{99.259} &
  91.852 &
  94.444 &
  77.778 &
  95.556 &
  97.407 &
  85.926 &
  96.296\; \\
FM &
  53.000 &
  57.000 &
  61.000 &
  55.000 &
  60.000 &
  62.000 &
  64.000 &
  63.000 &
  66.000 &
  67.000 &
  68.000 &
  67.000\; \\
HMD &
  18.919 &
  51.351 &
  48.649 &
  51.351 &
  36.486 &
  66.216 &
  66.216 &
  70.270 &
  \textbf{72.973} &
  66.216 &
  48.649 &
  \textbf{72.973}\; \\
IW &
  48.692 &
  38.524 &
  66.796 &
  67.052 &
  65.744 &
  18.996 &
  71.676 &
  72.424 &
  64.024 &
  \textbf{75.588} &
  10.012 &
  64.192\; \\
LIB &
  87.222 &
  90.556 &
  94.444 &
  93.333 &
  93.333 &
  70.556 &
  89.444 &
  34.444 &
  88.333 &
  90.000 &
  {\underline{95.556}} &
  88.333\; \\
LSST &
  55.109 &
  64.031 &
  66.383 &
  {\underline{66.504}} &
  64.274 &
  34.023 &
  41.525 &
  34.185 &
  44.120 &
  43.593 &
  51.014 &
  44.444\; \\
MI &
  50.000 &
  54.000 &
  50.000 &
  54.000 &
  51.000 &
  64.000 &
  68.000 &
  66.000 &
  68.000 &
  67.000 &
  \textbf{71.000} &
  67.000\; \\
NATO &
  88.333 &
  89.444 &
  91.111 &
  90.000 &
  93.333 &
  96.111 &
  97.222 &
  95.000 &
  98.333 &
  95.556 &
  \textbf{99.444} &
  96.111\; \\
PD &
  97.713 &
  98.228 &
  97.656 &
  97.856 &
  97.284 &
  92.481 &
  96.998 &
  46.941 &
  98.914 &
  98.513 &
  99.028 &
  98.599\; \\
PS &
  15.121 &
  27.826 &
  33.015 &
  {\underline{33.164}} &
  \textbf{33.642} &
  6.651 &
  16.224 &
  5.428 &
  14.524 &
  13.749 &
  4.921 &
  15.508\; \\
RS &
  80.263 &
  90.789 &
  90.132 &
  91.447 &
  90.132 &
  78.947 &
  81.579 &
  86.842 &
  89.474 &
  84.211 &
  \textbf{94.079} &
  91.447\; \\
SWJ &
  20.000 &
  46.667 &
  40.000 &
  40.000 &
  66.667 &
  53.333 &
  73.333 &
  66.667 &
  66.667 &
  {\underline{80.000}} &
  73.333 &
  73.333\; \\
\midrule
avg. acc (10) &
  70.354 &
  74.368 &
  78.156 &
  78.349 &
  77.600 &
  70.999 &
  72.760 &
  69.578 &
  75.168 &
  73.429 &
  77.443 &
  74.948\; \\
avg. acc &
  68.020 &
  73.306 &
  76.866 &
  77.124 &
  76.320 &
  68.247 &
  75.308 &
  69.001 &
  76.921 &
  76.938 &
  76.263 &
  77.279\; \\
avg. rank &
  17.00 &
  11.90 &
  9.50 &
  9.47 &
  10.10 &
  17.73 &
  12.57 &
  14.77 &
  9.33 &
  9.80 &
  8.40 &
  7.70\; \\
\# of top-1 &
  1 &
  3 &
  4 &
  {\underline{6}} &
  \textbf{8} &
  0 &
  1 &
  1 &
  3 &
  2 &
  {\underline{6}} &
  2\; \\
\# of top-5 &
  2 &
  10 &
  12 &
  13 &
  12 &
  0 &
  5 &
  3 &
  9 &
  8 &
  15 &
  11\; \\
\# of top-10 &
  4 &
  13 &
  19 &
  16 &
  16 &
  4 &
  9 &
  8 &
  15 &
  14 &
  20 &
  24\; \\
\midrule
\# of win/draw &
  30 &
  25 &
  21 &
  22 &
  23 &
  28 &
  24 &
  28 &
  22 &
  24 &
  21 &
  24\; \\
\# of lose/draw &
  1 &
  9 &
  12 &
  11 &
  9 &
  2 &
  8 &
  4 &
  10 &
  10 &
  12 &
  10\; \\
rank test &
  0.000 &
  0.000 &
  0.022 &
  0.024 &
  0.014 &
  0.000 &
  0.001 &
  0.000 &
  0.006 &
  0.015 &
  0.019 &
  0.005\; \\ \bottomrule
\end{tabular}
\footnotesize
All experiments were run on 1080 Ti with batch size 16 except below:
\begin{tablenotes}
\item[\tnote{8}] The experiments were run with batch size 8
\item[\tnote{4}] The experiments were run with batch size 4
\item[\tnote{A}] The experiments were run on A100 GPU
\end{tablenotes}
\end{threeparttable}
\addtolength{\tabcolsep}{+2pt}
}
\end{subtable}
\end{table}

\begin{table}[t]
\ContinuedFloat
\caption{\textit{(Continued.)} Classification accuracy\,(\%) of models on the 30 datasets of the UEA archive under the MambaSL protocol.
The best and the second-best are highlighted in bold and underline, respectively.}

\begin{subtable}{\columnwidth}
\caption{Classification results from FEDformer to RTSM.
``*'' indicates ``former'' in Transformer-based models.}
\resizebox{\columnwidth}{!}{
\addtolength{\tabcolsep}{-4pt}
\begin{threeparttable}
\begin{tabular}{lrrrrrrrrrr}
\toprule
 &
  \multicolumn{6}{c}{Transformer-based} &
  \multicolumn{1}{c}{\begin{tabular}[c]{@{}c@{}}Shape-\\based\end{tabular}} &
  \multicolumn{2}{c}{Mamba-based} &
  \multicolumn{1}{c}{\begin{tabular}[c]{@{}c@{}}Recursive-\\reasoning-\\based\end{tabular}} \\
  \cmidrule(r){2-7} \cmidrule(lr){8-8} \cmidrule(lr){9-10} \cmidrule(l){11-11}
\multicolumn{1}{c}{} &
  \multicolumn{1}{c}{\small{\begin{tabular}[c]{@{}c@{}}\;\,FED*\;\,\\ \citeyearpar{zhou2022fedformer}\end{tabular}}} &
  \multicolumn{1}{c}{\small{\begin{tabular}[c]{@{}c@{}}ETS*\\ \citeyearpar{woo2022etsformer}\end{tabular}}} &
  \multicolumn{1}{c}{\small{\begin{tabular}[c]{@{}c@{}}Cross*\\ \citeyearpar{zhang2023crossformer}\end{tabular}}} &
  \multicolumn{1}{c}{\footnotesize{\begin{tabular}[c]{@{}c@{}}PatchTST\\ \citeyearpar{yuqi2023patchtst}\end{tabular}}} &
  \multicolumn{1}{c}{\small{\begin{tabular}[c]{@{}c@{}}GPT4TS\\ \citeyearpar{zhou2023onefitsall}\end{tabular}}} &
  \multicolumn{1}{c}{\small{\begin{tabular}[c]{@{}c@{}}\;iTrans*\;\\ \citeyearpar{liu2024itransformer}\end{tabular}}} &
  \multicolumn{1}{c}{\footnotesize{\begin{tabular}[c]{@{}c@{}}InterpGN\\ \citeyearpar{wen2025interpgn}\end{tabular}}} &
  \multicolumn{1}{c}{\small{\begin{tabular}[c]{@{}c@{}}TSC\\Mamba\\ \citeyearpar{ahamed2025tscmamba}\end{tabular}}} &
  \multicolumn{1}{c}{\footnotesize{\begin{tabular}[c]{@{}c@{}}MambaSL\\ \citeyearpar{mambasl}\end{tabular}}} &
  \multicolumn{1}{c}{\footnotesize{\textbf{\begin{tabular}[c]{@{}c@{}}RTSM\\(ours)\end{tabular}}}} \\ \midrule
EC &
  33.840 &
  33.080 &
  43.346 &
  32.700 &
  32.319 &
  31.939 &
  28.897 &
  30.798 &
  42.586 &
  36.880\; \\
FD &
  69.892 &
  68.331 &
  69.637 &
  68.076 &
  68.615 &
  68.417 &
  63.791 &
  \textbf{70.204} &
  69.296 &
  69.550\; \\
HW &
  33.412 &
  35.647 &
  30.353 &
  29.529 &
  34.118 &
  31.294 &
  58.706 &
  45.059 &
  60.824 &
  \textbf{63.530}\; \\
HB &
  78.049 &
  78.537 &
  79.024 &
  73.659 &
  79.024 &
  78.049 &
  80.488 &
  78.537 &
  80.488 &
  \underline{80.980}\; \\
JV &
  97.568 &
  98.378 &
  98.378 &
  97.027 &
  98.108 &
  98.108 &
  \textbf{99.730} &
  98.378 &
  98.649 &
  \textbf{99.730}\; \\
PEMS &
  89.017 &
  87.861 &
  89.595\tnote{8} &
  89.017\tnote{4} &
  88.439 &
  91.329\tnote{8} &
  86.127 &
  90.173 &
  85.549 &
  91.910\; \\
SRS1 &
  75.085 &
  92.150 &
  \underline{93.857} &
  89.078 &
  93.174 &
  \underline{93.857} &
  90.444 &
  92.833 &
  92.491 &
  90.780\; \\
SRS2 &
  59.444 &
  60.000 &
  61.667 &
  61.667 &
  60.556 &
  58.889 &
  58.333 &
  62.222 &
  \textbf{65.000} &
  61.670\; \\
SAD &
  99.454 &
  99.000 &
  98.863 &
  98.818 &
  99.591 &
  99.318 &
  \underline{99.864} &
  95.816 &
  \textbf{99.955} &
  99.500\; \\
UW &
  79.688 &
  89.688 &
  88.438 &
  89.375 &
  89.063 &
  90.000 &
  88.438 &
  \textbf{95.313} &
  93.438 &
  91.870\; \\
AWR &
  92.333 &
  98.667 &
  98.667 &
  98.667 &
  98.667 &
  \underline{99.000} &
  \underline{99.000} &
  \textbf{99.333} &
  \textbf{99.333} &
  \textbf{99.330}\; \\
AF &
  \underline{73.333} &
  \textbf{80.000} &
  \underline{73.333} &
  \textbf{80.000} &
  66.667 &
  66.667 &
  46.667 &
  66.667 &
  53.333 &
  60.000\; \\
BM &
  95.000 &
  \textbf{100.000} &
  \textbf{100.000} &
  85.000 &
  \textbf{100.000} &
  \underline{97.500} &
  \textbf{100.000} &
  \textbf{100.000} &
  \textbf{100.000} &
  \textbf{100.000}\; \\
CT &
  97.284 &
  98.468 &
  98.955 &
  98.398 &
  99.373 &
  98.886 &
  \textbf{99.791} &
  99.513 &
  \underline{99.721} &
  \underline{99.720}\; \\
CR &
  90.278 &
  \underline{98.611} &
  97.222 &
  97.222 &
  97.222 &
  97.222 &
  \textbf{100.000} &
  \textbf{100.000} &
  \textbf{100.000} &
  \textbf{100.000}\; \\
DDG &
  42.000 &
  42.000 &
  72.000\tnote{4} &
  40.000\tnote{4} &
  66.000 &
  60.000\tnote{8} &
  \underline{76.000}\tnote{A} &
  50.000 &
  70.000 &
  \textbf{78.000}\; \\
EW &
  63.359\tnote{4} &
  70.992\tnote{4} &
  58.015\tnote{4} &
  57.252 &
  66.412 &
  59.542 &
  82.443\tnote{A} &
  88.550 &
  83.969\tnote{4} &
  90.840\; \\
EP &
  92.754 &
  92.754 &
  95.652 &
  98.551 &
  89.855 &
  80.435 &
  97.826 &
  \underline{99.275} &
  97.826 &
  \underline{99.280}\; \\
ER &
  95.926 &
  96.296 &
  97.407 &
  96.667 &
  96.296 &
  94.815 &
  87.778 &
  97.037 &
  93.704 &
  93.700\; \\
FM &
  \textbf{72.000} &
  66.000 &
  64.000 &
  63.000 &
  66.000 &
  64.000 &
  67.000 &
  66.000 &
  \underline{71.000} &
  67.000\; \\
HMD &
  \underline{71.622} &
  63.514 &
  66.216 &
  63.514 &
  70.270 &
  58.108 &
  45.946 &
  60.811 &
  70.270 &
  63.510\; \\
IW &
  65.244 &
  64.868 &
  \underline{73.712} &
  60.936 &
  61.396 &
  73.436 &
  67.180 &
  68.244\tnote{*} &
  66.304 &
  66.400\; \\
LIB &
  93.333 &
  90.000 &
  90.556 &
  81.667 &
  87.778 &
  91.667 &
  \textbf{97.222} &
  91.667 &
  91.667 &
  93.890\; \\
LSST &
  40.998 &
  41.241 &
  41.768 &
  56.732 &
  38.078 &
  40.673 &
  59.570 &
  60.057 &
  45.580 &
  \textbf{70.110}\; \\
MI &
  65.000 &
  67.000\tnote{8} &
  68.000\tnote{4} &
  \underline{69.000} &
  67.000 &
  67.000 &
  67.000\tnote{A} &
  67.000 &
  \underline{69.000} &
  67.000\; \\
NATO &
  96.111 &
  93.333 &
  93.889 &
  80.556 &
  96.111 &
  88.889 &
  \underline{98.889} &
  92.222 &
  \underline{98.889} &
  98.330\; \\
PD &
  98.599 &
  98.542 &
  98.513 &
  98.485 &
  98.513 &
  98.428 &
  \underline{99.342} &
  98.513 &
  99.257 &
  \textbf{99.370}\; \\
PS &
  11.691 &
  14.316 &
  14.256 &
  15.091 &
  13.123 &
  11.542 &
  33.075 &
  23.800 &
  30.331 &
  26.250\; \\
RS &
  84.868 &
  85.526 &
  87.500 &
  85.526 &
  86.842 &
  81.579 &
  91.447 &
  90.789 &
  \underline{92.763} &
  92.110\; \\
SWJ &
  66.667 &
  66.667\tnote{8} &
  \underline{80.000} &
  \textbf{86.667} &
  73.333 &
  73.333 &
  60.000 &
  73.333 &
  73.333 &
  73.330\; \\
\midrule
avg. acc (10) &
  71.545 &
  74.267 &
  75.316 &
  72.895 &
  74.301 &
  74.120 &
  75.482 &
  75.933 &
  \textbf{78.827} &
  \underline{78.640}\; \\
avg. acc &
  74.128 &
  75.716 &
  77.427 &
  74.729 &
  76.065 &
  74.797 &
  77.700 &
  78.405 &
  \underline{79.819} &
  \textbf{80.819}\; \\
avg. rank &
  13.17 &
  11.43 &
  9.13 &
  13.40 &
  10.23 &
  12.63 &
  8.50 &
  7.67 &
  \underline{5.73} &
  \textbf{5.07}\; \\
\# of top-1 &
  1 &
  2 &
  1 &
  2 &
  1 &
  0 &
  5 &
  5 &
  5 &
  \textbf{8}\; \\
\# of top-5 &
  5 &
  2 &
  11 &
  4 &
  6 &
  4 &
  13 &
  10 &
  \underline{17} &
  \textbf{19}\; \\
\# of top-10 &
  10 &
  12 &
  16 &
  8 &
  15 &
  9 &
  20 &
  \underline{25} &
  \underline{25} &
  \textbf{26}\; \\
\midrule
\# of win/draw &
  25 &
  27 &
  21 &
  26 &
  25 &
  26 &
  24 &
  23 &
  19 &
  \; \\
\# of lose/draw &
  5 &
  6 &
  11 &
  6 &
  8 &
  6 &
  11 &
  13 &
  17 &
  \; \\
rank test &
  0.000 &
  0.000 &
  0.069 &
  0.001 &
  0.001 &
  0.001 &
  0.002 &
  0.020 &
  0.355 &
  \; \\ \bottomrule
\end{tabular}
\footnotesize
The MambaSL benchmark experiments were run on 1080 Ti with batch size 16 except below; RTSM was run on NVIDIA GH200.
\begin{tablenotes}
\item[\tnote{8}] The experiments were run with batch size 8
\item[\tnote{4}] The experiments were run with batch size 4
\item[\tnote{A}] The experiments were run on A100 GPU
\item[\tnote{*}] Little hyperparameter search was performed since it took over two weeks conducting a single experiment.
\end{tablenotes}
\end{threeparttable}
\addtolength{\tabcolsep}{+4pt}
}
\end{subtable}
\end{table}

\clearpage
\subsection{Full Results on the UEA Benchmark under Validation protocol}
Table~\ref{tab:full-uea30-result-validation} reports the complete per-dataset results for the evaluated deep-learning models under the validation protocol. Model and checkpoint selection are performed exclusively on the validation split, with the official test set reserved for final evaluation. Full protocol and implementation details are provided in Appendix~\ref{appendix:experimental_setup}.
\begin{table}[!htbp]
\centering
\captionsetup{font=footnotesize,skip=2pt}
\caption{Classification accuracy (\%) of deep-learning models on the 30 UEA datasets under the validation protocol. Best and second-best results are shown in bold and underlined, respectively.}
\label{tab:full-uea30-result-validation}

\renewcommand{\arraystretch}{0.78}
\resizebox{0.95\columnwidth}{!}{%
\begin{threeparttable}
\begin{tabular}{lrrrrrrr}
\toprule
& \multicolumn{3}{c}{Linear/MLP} & \multicolumn{4}{c}{Convolutional/spectral/multiscale} \\
\cmidrule(lr){2-4}\cmidrule(lr){5-8}
&
\begin{tabular}[c]{@{}c@{}}DLinear\\ \citeyearpar{zeng2023dlinear}\end{tabular} &
\begin{tabular}[c]{@{}c@{}}LightTS\\ \citeyearpar{tianping2022lightts}\end{tabular} &
\begin{tabular}[c]{@{}c@{}}MTS-\\Mixers\\ \citeyearpar{li2023mtsmixer}\end{tabular} &
\begin{tabular}[c]{@{}c@{}}TimesNet\\ \citeyearpar{wu2023timesnet}\end{tabular} &
\begin{tabular}[c]{@{}c@{}}Modern\\TCN\\ \citeyearpar{luo2024moderntcn}\end{tabular} &
\begin{tabular}[c]{@{}c@{}}TSLANet\\ \citeyearpar{eldele2024tslanet}\end{tabular} &
\begin{tabular}[c]{@{}c@{}}Time\\Mixer++\\ \citeyearpar{wang2025timemixerpp}\end{tabular} \\
\midrule

EC   & 25.480 & 28.140 & 22.050 & 28.520 & 23.950 & \textbf{30.420} & 24.330 \\
FD   & 67.820 & 64.560 & 65.580 & 66.030 & 64.100 & 65.490 & 64.390 \\
HW   & 21.530 & 19.530 & 14.940 & 27.760 & 19.650 & \textbf{56.350} & 26.000 \\
HB   & 73.170 & 73.170 & 73.660 & \textbf{77.070} & 68.780 & \textbf{77.070} & 75.610 \\
JV   & 95.680 & 92.970 & 92.970 & 97.840 & 96.220 & 96.490 & 97.030 \\
PEMS & 72.250 & 74.570 & 72.830 & 76.880 & 81.500 & 80.350 & 80.350 \\
SRS1 & \underline{87.370} & 81.230 & 77.470 & 82.940 & 86.690 & 67.240 & 79.520 \\
SRS2 & 50.000 & 53.890 & 48.890 & 49.440 & 52.780 & 51.110 & 52.780 \\
SAD  & 95.180 & 98.180 & 98.680 & 99.140 & 97.140 & \underline{99.320} & 99.140 \\
UW   & 80.940 & 81.250 & 79.370 & 84.690 & 84.690 & \textbf{90.310} & 83.750 \\
AWR  & 96.330 & 95.330 & 95.000 & 97.670 & 98.000 & 97.670 & \underline{98.330} \\
AF   & 33.330 & 33.330 & 20.000 & \underline{40.000} & 20.000 & 33.330 & 26.670 \\
BM   & 85.000 & \textbf{100.000} & 90.000 & \underline{97.500} & 95.000 & \textbf{100.000} & \underline{97.500} \\
CT   & 97.080 & 98.470 & 97.560 & 97.350 & 98.120 & 98.470 & 98.400 \\
CR   & 90.280 & 88.890 & 93.060 & 90.280 & 93.060 & 97.220 & 94.440 \\
DDG  & 56.000 & 48.000 & 48.000 & 48.000 & 44.000 & 50.000 & 52.000 \\
EW   & 28.240 & 47.330 & 49.620 & 51.910 & 50.380 & 70.230 & 51.150 \\
EP   & 56.520 & 91.300 & 62.320 & 89.860 & 92.030 & \textbf{98.550} & 92.030 \\
ER   & 92.220 & \underline{94.440} & 68.890 & 93.330 & \textbf{94.810} & 72.960 & 92.960 \\
FM   & 57.000 & 53.000 & 52.000 & 56.000 & \textbf{62.000} & 60.000 & 59.000 \\
HMD  & 54.050 & 37.840 & \underline{63.510} & \textbf{64.860} & 33.780 & 27.030 & 58.110 \\
IW   & 18.280 & 70.330 & 70.700 & 62.280 & \textbf{74.280} & 10.080 & 61.250 \\
LIB  & 68.890 & 79.440 & 29.440 & 82.220 & 85.000 & \underline{91.110} & 77.220 \\
LSST & 32.850 & 20.560 & 8.960 & 33.290 & 36.540 & 56.810 & 35.850 \\
MI   & \underline{59.000} & 50.000 & 51.000 & \textbf{60.000} & 41.000 & 51.000 & 49.000 \\
NATO & 93.890 & 93.330 & 92.780 & 93.330 & 90.560 & \textbf{97.780} & 90.560 \\
PD   & 92.140 & 96.430 & 44.170 & 98.310 & 97.740 & 98.200 & 97.970 \\
PS   & 6.470 & 14.460 & 4.290 & 11.870 & 11.180 & 9.870 & 13.030 \\
RS   & 72.370 & 73.030 & 83.550 & 84.210 & 76.970 & \underline{90.790} & 80.260 \\
SWJ  & \underline{46.670} & 33.330 & 26.670 & 33.330 & 26.670 & \textbf{53.330} & 40.000 \\
\midrule
avg. acc (10) & 66.942 & 66.749 & 64.644 & 69.031 & 67.550 & \underline{71.415} & 68.290 \\
avg. acc      & 63.534 & 66.211 & 59.932 & 69.197 & 66.554 & 69.286 & 68.288 \\
avg. rank     & 11.13 & 10.03 & 12.30 & 7.03 & 9.27 & 6.07 & 7.90 \\
\# top-1       & 0 & 1 & 0 & 3 & 3 & \textbf{8} & 0 \\
\# top-5       & 6 & 7 & 2 & 9 & 7 & 17 & 10 \\
\# top-10      & 12 & 14 & 11 & 23 & 19 & \textbf{25} & 22 \\
\midrule
\# win/draw  & 22 & 24 & 26 & 21 & 25 & 19 & 24 \\
\# lose/draw & 8 & 7 & 5 & 9 & 7 & 14 & 8 \\
rank test    & 0.002 & 0.003 & 0.000 & 0.029 & 0.001 & 0.162 & 0.003 \\
\bottomrule
\end{tabular}
\begin{tablenotes}[flushleft]
\scriptsize
\item All deep-learning baselines and RTSM were run on NVIDIA GH200 GPUs.
\end{tablenotes}
\end{threeparttable}%
}
\vspace{-0.4em}
\end{table}

\begin{table}[t]
\ContinuedFloat
\caption{\textit{(Continued.)} Classification accuracy\,(\%) of deep-learning models on the 30 datasets of the UEA archive under the validation protocol.
The best and the second-best are highlighted in bold and underline, respectively.}

\begin{subtable}{\columnwidth}
\caption{Classification results from FEDformer to RTSM.
``*'' indicates ``former'' in Transformer-based models.}
\resizebox{\columnwidth}{!}{
\addtolength{\tabcolsep}{-4pt}
\begin{threeparttable}
\begin{tabular}{lrrrrrrrrrr}
\toprule
 &
  \multicolumn{6}{c}{Transformer-based} &
  \multicolumn{1}{c}{\begin{tabular}[c]{@{}c@{}}Shape-\\based\end{tabular}} &
  \multicolumn{2}{c}{Mamba-based} &
  \multicolumn{1}{c}{\begin{tabular}[c]{@{}c@{}}Recursive-\\reasoning-\\based\end{tabular}} \\
  \cmidrule(r){2-7} \cmidrule(lr){8-8} \cmidrule(lr){9-10} \cmidrule(l){11-11}
\multicolumn{1}{c}{} &
  \multicolumn{1}{c}{\small{\begin{tabular}[c]{@{}c@{}}\;\,FED*\;\,\\ \citeyearpar{zhou2022fedformer}\end{tabular}}} &
  \multicolumn{1}{c}{\small{\begin{tabular}[c]{@{}c@{}}ETS*\\ \citeyearpar{woo2022etsformer}\end{tabular}}} &
  \multicolumn{1}{c}{\small{\begin{tabular}[c]{@{}c@{}}Cross*\\ \citeyearpar{zhang2023crossformer}\end{tabular}}} &
  \multicolumn{1}{c}{\footnotesize{\begin{tabular}[c]{@{}c@{}}PatchTST\\ \citeyearpar{yuqi2023patchtst}\end{tabular}}} &
  \multicolumn{1}{c}{\small{\begin{tabular}[c]{@{}c@{}}GPT4TS\\ \citeyearpar{zhou2023onefitsall}\end{tabular}}} &
  \multicolumn{1}{c}{\small{\begin{tabular}[c]{@{}c@{}}\;iTrans*\;\\ \citeyearpar{liu2024itransformer}\end{tabular}}} &
  \multicolumn{1}{c}{\footnotesize{\begin{tabular}[c]{@{}c@{}}InterpGN\\ \citeyearpar{wen2025interpgn}\end{tabular}}} &
  \multicolumn{1}{c}{\small{\begin{tabular}[c]{@{}c@{}}TSC\\Mamba\\ \citeyearpar{ahamed2025tscmamba}\end{tabular}}} &
  \multicolumn{1}{c}{\footnotesize{\begin{tabular}[c]{@{}c@{}}MambaSL\\ \citeyearpar{mambasl}\end{tabular}}} &
  \multicolumn{1}{c}{\footnotesize{\textbf{\begin{tabular}[c]{@{}c@{}}RTSM\\(ours)\end{tabular}}}} \\ \midrule
EC &
  25.860 &
  23.570 &
  \textbf{30.420} &
  26.620 &
  23.570 &
  27.380 &
  25.480 &
  25.100 &
  27.000 &
  \underline{30.418}\; \\
FD &
  \underline{68.620} &
  65.270 &
  67.170 &
  63.710 &
  65.010 &
  64.780 &
  62.150 &
  \textbf{69.100} &
  67.110 &
  67.196\; \\
HW &
  24.820 &
  28.710 &
  22.590 &
  20.940 &
  26.710 &
  23.410 &
  \underline{50.820} &
  26.470 &
  50.350 &
  47.412\; \\
HB &
  70.240 &
  70.730 &
  73.660 &
  60.980 &
  75.120 &
  72.200 &
  45.850 &
  \underline{76.100} &
  69.760 &
  68.780\; \\
JV &
  96.220 &
  96.220 &
  95.680 &
  93.780 &
  97.300 &
  90.540 &
  \textbf{98.380} &
  97.030 &
  96.760 &
  \underline{98.108}\; \\
PEMS &
  81.500 &
  77.460 &
  \underline{87.280} &
  83.820 &
  75.140 &
  82.660 &
  80.920 &
  83.240 &
  72.830 &
  \textbf{87.861}\; \\
SRS1 &
  69.970 &
  80.890 &
  81.230 &
  82.250 &
  86.010 &
  \textbf{89.760} &
  72.700 &
  85.670 &
  73.380 &
  82.253\; \\
SRS2 &
  51.670 &
  52.220 &
  47.220 &
  \underline{55.560} &
  48.330 &
  50.560 &
  \textbf{56.110} &
  50.560 &
  52.780 &
  51.667\; \\
SAD &
  97.950 &
  98.500 &
  98.320 &
  97.770 &
  98.000 &
  98.500 &
  99.230 &
  98.820 &
  \textbf{99.590} &
  98.363\; \\
UW &
  75.620 &
  85.000 &
  82.810 &
  85.620 &
  83.440 &
  84.380 &
  85.000 &
  86.560 &
  \underline{88.440} &
  85.938\; \\
AWR &
  88.000 &
  97.330 &
  98.000 &
  95.330 &
  97.670 &
  97.000 &
  \underline{98.330} &
  96.670 &
  98.000 &
  \textbf{98.667}\; \\
AF &
  \textbf{46.670} &
  20.000 &
  \underline{40.000} &
  \underline{40.000} &
  33.330 &
  26.670 &
  20.000 &
  \underline{40.000} &
  33.330 &
  26.667\; \\
BM &
  95.000 &
  87.500 &
  92.500 &
  52.500 &
  90.000 &
  90.000 &
  \textbf{100.000} &
  95.000 &
  \textbf{100.000} &
  \textbf{100.000}\; \\
CT &
  96.660 &
  97.350 &
  96.730 &
  97.280 &
  98.540 &
  97.630 &
  \textbf{99.370} &
  98.190 &
  98.820 &
  \underline{99.304}\; \\
CR &
  86.110 &
  94.440 &
  84.720 &
  91.670 &
  95.830 &
  87.500 &
  97.220 &
  95.830 &
  \textbf{98.610} &
  \underline{97.222}\; \\
DDG &
  22.000 &
  16.000 &
  58.000 &
  20.000 &
  50.000 &
  52.000 &
  \textbf{64.000} &
  40.000 &
  52.000 &
  \underline{62.000}\; \\
EW &
  47.330 &
  56.490 &
  46.560 &
  45.800 &
  51.910 &
  44.270 &
  19.080 &
  \underline{78.630} &
  71.760 &
  \textbf{84.733}\; \\
EP &
  88.410 &
  90.580 &
  94.200 &
  94.200 &
  81.160 &
  73.910 &
  \underline{97.830} &
  94.200 &
  95.650 &
  94.928\; \\
ER &
  91.850 &
  92.960 &
  93.700 &
  92.220 &
  90.740 &
  89.630 &
  87.780 &
  91.850 &
  87.780 &
  88.519\; \\
FM &
  58.000 &
  51.000 &
  55.000 &
  51.000 &
  56.000 &
  59.000 &
  59.000 &
  \underline{61.000} &
  55.000 &
  55.000\; \\
HMD &
  58.110 &
  44.590 &
  41.890 &
  41.890 &
  41.890 &
  32.430 &
  37.840 &
  56.760 &
  58.110 &
  58.108\; \\
IW &
  62.780 &
  64.030 &
  72.320 &
  60.290 &
  59.260 &
  \underline{72.360} &
  64.100 &
  68.980 &
  64.110 &
  64.472\; \\
LIB &
  83.890 &
  84.440 &
  82.220 &
  77.780 &
  73.890 &
  87.780 &
  \textbf{95.000} &
  83.890 &
  82.220 &
  87.222\; \\
LSST &
  20.920 &
  36.620 &
  36.580 &
  52.070 &
  9.980 &
  11.030 &
  56.450 &
  \underline{57.540} &
  40.880 &
  \textbf{65.369}\; \\
MI &
  48.000 &
  55.000 &
  58.000 &
  52.000 &
  53.000 &
  48.000 &
  48.000 &
  51.000 &
  49.000 &
  54.000\; \\
NATO &
  87.220 &
  88.330 &
  86.670 &
  77.220 &
  92.780 &
  82.780 &
  95.000 &
  89.440 &
  \underline{95.560} &
  95.556\; \\
PD &
  98.230 &
  97.630 &
  97.030 &
  96.970 &
  98.060 &
  97.400 &
  \underline{98.340} &
  97.570 &
  97.570 &
  \textbf{98.799}\; \\
PS &
  10.410 &
  11.300 &
  10.680 &
  12.320 &
  11.060 &
  9.870 &
  \textbf{30.810} &
  19.890 &
  \underline{28.390} &
  24.217\; \\
RS &
  76.970 &
  76.970 &
  82.890 &
  76.970 &
  84.870 &
  76.970 &
  88.160 &
  85.530 &
  \textbf{91.450} &
  84.868\; \\
SWJ &
  20.000 &
  40.000 &
  26.670 &
  33.330 &
  40.000 &
  20.000 &
  \underline{46.670} &
  33.330 &
  40.000 &
  26.667\; \\
\midrule
avg. acc (10) &
  66.247 &
  67.857 &
  68.638 &
  67.105 &
  67.863 &
  68.417 &
  67.664 &
  69.865 &
  69.800 &
  \textbf{71.800}\; \\
avg. acc &
  64.968 &
  66.038 &
  68.025 &
  64.396 &
  66.287 &
  64.680 &
  69.321 &
  71.132 &
  \underline{71.208} &
  \textbf{72.810}\; \\
avg. rank &
  10.77 &
  9.57 &
  8.90 &
  10.73 &
  9.27 &
  10.67 &
  6.37 &
  6.43 &
  \underline{6.00} &
  \textbf{4.87}\; \\
\# of top-1 &
  1 &
  0 &
  1 &
  0 &
  0 &
  1 &
  \underline{7} &
  1 &
  4 &
  \underline{7}\; \\
\# of top-5 &
  4 &
  5 &
  10 &
  6 &
  7 &
  6 &
  \underline{18} &
  14 &
  16 &
  \textbf{21}\; \\
\# of top-10 &
  13 &
  18 &
  16 &
  13 &
  17 &
  11 &
  22 &
  \textbf{25} &
  \underline{24} &
  \textbf{25}\; \\
\midrule
\# of win/draw &
  25 &
  24 &
  25 &
  26 &
  24 &
  23 &
  18 &
  19 &
  19 &
  \; \\
\# of lose/draw &
  7 &
  6 &
  8 &
  5 &
  7 &
  8 &
  14 &
  11 &
  15 &
  \; \\
rank test &
  0.000 &
  0.000 &
  0.006 &
  0.000 &
  0.002 &
  0.000 &
  0.123 &
  0.087 &
  0.238 &
  \; \\
\bottomrule
\end{tabular}
\footnotesize
All deep-learning baselines and RTSM were run on NVIDIA GH200 GPUs.
\end{threeparttable}
\addtolength{\tabcolsep}{+4pt}
}
\end{subtable}
\end{table}

\clearpage
\subsection{Recursive Architecture and Training Ablations}
\label{appendix:trm-analysis-result}

Configuration names encode the corresponding architectural and training
choices. \texttt{H} denotes the number of outer recursive cycles,
\texttt{L} the number of inner cycles used by two-state variants,
and \texttt{G} the number of H-cycles through which gradients are propagated;
accordingly, \texttt{2G}/\texttt{3G} extend the gradient window.
\texttt{stepsN} denotes the maximum number of deep-supervision steps.
For two-state models, \texttt{shared} uses a single reasoning network $f$ for
both state updates, as in \gls{trm}~\citep{jolicoeur2025trm}, whereas
\texttt{separate} uses distinct $f_H/f_L$ networks, implementing the network
separation used by \gls{hrm}~\citep{wang2025hrm}. All other training choices follow
the \gls{trm} recipe, including its gradient schedule, loss, halting, and parameter
EMA; hence, the separate-network variant is not an evaluation of \gls{hrm} itself.
\texttt{Conv} denotes the simple convolutional patch tokenizer, which uses
strided 1-D convolutions to map local temporal windows into patch
representations; configurations
without this prefix use the proposed time-series tokenizer.

Additional suffixes denote further architectural or training modifications.
\texttt{tokenizer\_only} omits the attention-based refinement stage and applies the
sequence-level prediction head directly to the tokenizer representation.
\texttt{nocarry} resets the latent state between supervision steps. \texttt{forcefull} disables
Q-based halting and forces every sample through the full training-step budget.
\texttt{mlpt} replaces attention with a temporal \gls{mlp}.
\texttt{meanpooling} and \texttt{adaptivepooling\_noQ} replace the default pooling
operation. The latter applies learned weighted pooling only to the classification head, whereas the reference model applies learned weighted pooling to both the classification and halting heads. \texttt{rima} enables the learned EMA-style latent-state reweighting
of RIMA~\citep{komisarczyk2026recursiveinferencemachinesneural}, while \texttt{pdc} applies the Progressive Depth
Curriculum of CGAR~\citep{qasim2025acceleratingtrainingspeedtiny}, gradually increasing recursive depth during
training.

\begin{table}[H]
\centering
\caption{Classification accuracy (\%) across the recursive architecture and training ablations on eight UEA datasets (CT, DDG, HB, JV, LSST, NATO, PD, and SAD) under validation-based selection. Each seed's score is the equally weighted mean across these datasets. Entries report the mean $\pm$ sample standard deviation of the three scores for seeds 0, 1, and 2. Bold marks the reference configuration.}
\label{tab:trm-analysis-results}
\begin{tabular}{lc}
\hline
\textbf{Configuration} & \textbf{Avg. Acc. (\%)} \\
\hline
\multicolumn{2}{l}{\textit{Main-text ablations}} \\
\texttt{Conv\_tokenizer\_only}              & $73.72 \pm 0.71$ \\
\texttt{Conv\_3H6L1G\_steps16\_shared}   & $78.96 \pm 1.03$ \\
\texttt{Conv\_3H6L1G\_steps16\_separate} & $79.69 \pm 0.62$ \\
\texttt{tokenizer\_only}                 & $84.79 \pm 0.41$ \\
\texttt{3H6L1G\_steps16\_shared}         & $84.98 \pm 1.04$ \\
\texttt{3H6L1G\_steps16\_separate}       & $84.29 \pm 1.26$ \\
\texttt{3H1G\_steps1}                       & $85.98 \pm 0.71$ \\
\texttt{3H1G\_steps16\_forcefull}           & $85.05 \pm 0.91$ \\
\texttt{1H1G\_steps16}                      & $85.77 \pm 0.59$ \\
\texttt{2H1G\_steps16}                      & $85.60 \pm 0.42$ \\
\texttt{3H1G\_steps16}                      & $\mathbf{86.10 \pm 0.31}$ \\
\texttt{4H1G\_steps16}                      & $85.61 \pm 0.75$ \\
\texttt{5H1G\_steps16}                      & $85.78 \pm 0.64$ \\
\hline
\multicolumn{2}{l}{\textit{Additional ablations}} \\
\texttt{1H1G\_steps16\_nocarry}               & $86.00 \pm 0.76$ \\
\texttt{3H1G\_steps16\_nocarry}             & $86.09 \pm 0.59$ \\
\texttt{3H2G\_steps16}                      & $86.06 \pm 0.63$ \\
\texttt{3H3G\_steps16}                      & $85.91 \pm 0.33$ \\
\texttt{3H1G\_steps16\_meanpooling}         & $86.14 \pm 0.33$ \\
\texttt{3H1G\_steps16\_adaptivepooling\_noQ}     & $85.85 \pm 0.20$ \\
\texttt{3H1G\_steps16\_mlpt}                & $83.85 \pm 0.47$ \\
\texttt{3H1G\_steps16\_rima}                & $86.00 \pm 0.55$ \\
\texttt{3H1G\_steps16\_pdc}                 & $86.02 \pm 0.28$ \\
\texttt{3H1G\_steps16\_pdc\_rima}           & $85.96 \pm 0.87$ \\
\hline
\end{tabular}
\end{table}

\begin{table}[H]
\centering
\caption{
Comparison of the proposed \texttt{3H1G\_steps16} configuration with the
\texttt{tokenizer\_only} baseline across the eight ablation datasets.
Results report classification accuracy (\%) for each seed and the mean
$\pm$ sample standard deviation across three seeds. In the Overall row, datasets are averaged equally within each seed before computing the mean and standard deviation. $\Delta$ denotes the difference
in mean accuracy between the proposed configuration and the tokenizer-only
baseline.
}
\label{tab:recursive-vs-tokenizer}
\resizebox{\linewidth}{!}{
\begin{tabular}{lccccccccr}
\hline
& \multicolumn{4}{c}{\textbf{\texttt{3H1G\_steps16}}}
& \multicolumn{4}{c}{\textbf{\texttt{tokenizer\_only}}}
& \\
\cline{2-5}\cline{6-9}
\textbf{Dataset}
& \textbf{Seed 0}
& \textbf{Seed 1}
& \textbf{Seed 2}
& \textbf{Mean $\pm$ Std.}
& \textbf{Seed 0}
& \textbf{Seed 1}
& \textbf{Seed 2}
& \textbf{Mean $\pm$ Std.}
& \textbf{$\Delta$} \\
\hline

CharacterTrajectories
& 99.234 & 99.095 & 98.886
& $\mathbf{99.072 \pm 0.175}$
& 95.891 & 96.031 & 96.379
& $96.100 \pm 0.251$
& +2.971 \\

PenDigits
& 98.571 & 98.742 & 99.085
& $\mathbf{98.799 \pm 0.262}$
& 98.485 & 98.113 & 98.228
& $98.275 \pm 0.190$
& +0.524 \\

SpokenArabicDigits
& 98.408 & 98.681 & 98.590
& $\mathbf{98.560 \pm 0.139}$
& 96.362 & 96.771 & 97.317
& $96.817 \pm 0.479$
& +1.743 \\

JapaneseVowels
& 97.297 & 97.838 & 96.757
& $\mathbf{97.297 \pm 0.541}$
& 96.757 & 97.027 & 96.757
& $96.847 \pm 0.156$
& +0.450 \\

NATOPS
& 91.111 & 89.444 & 91.111
& $90.555 \pm 0.962$
& 89.444 & 92.778 & 93.333
& $\mathbf{91.852 \pm 2.103}$
& -1.296 \\

Heartbeat
& 73.659 & 75.122 & 74.146
& $\mathbf{74.309 \pm 0.745}$
& 73.659 & 77.561 & 71.707
& $\mathbf{74.309 \pm 2.981}$
& +0.000 \\

LSST
& 67.680 & 66.951 & 67.883
& $\mathbf{67.505 \pm 0.490}$
& 66.869 & 66.504 & 64.882
& $66.085 \pm 1.058$
& +1.419 \\

DuckDuckGeese
& 60.000 & 64.000 & 64.000
& $\mathbf{62.667 \pm 2.309}$
& 62.000 & 56.000 & 56.000
& $58.000 \pm 3.464$
& +4.667 \\

\hline
\textbf{Overall}
& 85.745 & 86.234 & 86.307
& $\mathbf{86.095 \pm 0.306}$
& 84.933 & 85.098 & 84.325
& $84.786 \pm 0.407$
& \textbf{+1.310} \\
\hline

\end{tabular}
}
\end{table}

\begin{table}[H]
\centering
\caption{
Comparison of the proposed \texttt{3H1G\_steps16} configuration with the
\texttt{3H6L1G\_steps16\_shared} two-state variant across the eight ablation datasets.
Results report classification accuracy (\%) for each seed and the mean
$\pm$ sample standard deviation across three seeds. In the Overall row, datasets are averaged equally within each seed before computing the mean and standard deviation. $\Delta$ denotes the difference
in mean accuracy between the proposed configuration and the shared-network two-state variant.
}
\label{tab:singlestate-vs-trm}
\resizebox{\linewidth}{!}{
\begin{tabular}{lccccccccr}
\hline
& \multicolumn{4}{c}{\textbf{\texttt{3H1G\_steps16}}}
& \multicolumn{4}{c}{\textbf{\texttt{3H6L1G\_steps16\_shared}}}
& \\
\cline{2-5}\cline{6-9}
\textbf{Dataset}
& \textbf{Seed 0}
& \textbf{Seed 1}
& \textbf{Seed 2}
& \textbf{Mean $\pm$ Std.}
& \textbf{Seed 0}
& \textbf{Seed 1}
& \textbf{Seed 2}
& \textbf{Mean $\pm$ Std.}
& \textbf{$\Delta$} \\
\hline

CharacterTrajectories
& 99.234 & 99.095 & 98.886
& $\mathbf{99.072 \pm 0.175}$
& 99.304 & 98.189 & 98.886
& $98.793 \pm 0.563$
& +0.279 \\

PenDigits
& 98.571 & 98.742 & 99.085
& $\mathbf{98.799 \pm 0.262}$
& 98.856 & 98.571 & 98.370
& $98.599 \pm 0.244$
& +0.200 \\

SpokenArabicDigits
& 98.408 & 98.681 & 98.590
& $\mathbf{98.560 \pm 0.139}$
& 97.954 & 98.090 & 98.363
& $98.136 \pm 0.208$
& +0.424 \\

JapaneseVowels
& 97.297 & 97.838 & 96.757
& $\mathbf{97.297 \pm 0.541}$
& 98.108 & 96.486 & 96.486
& $97.027 \pm 0.936$
& +0.271 \\

NATOPS
& 91.111 & 89.444 & 91.111
& $\mathbf{90.555 \pm 0.962}$
& 82.778 & 77.778 & 95.556
& $85.371 \pm 9.168$
& +5.185 \\

Heartbeat
& 73.659 & 75.122 & 74.146
& $\mathbf{74.309 \pm 0.745}$
& 75.122 & 70.244 & 72.195
& $72.520 \pm 2.455$
& +1.789 \\

LSST
& 67.680 & 66.951 & 67.883
& $\mathbf{67.505 \pm 0.490}$
& 66.464 & 65.815 & 67.843
& $66.707 \pm 1.036$
& +0.797 \\

DuckDuckGeese
& 60.000 & 64.000 & 64.000
& $\mathbf{62.667 \pm 2.309}$
& 62.000 & 66.000 & 60.000
& $\mathbf{62.667 \pm 3.055}$
& +0.000 \\

\hline
\textbf{Overall}
& 85.745 & 86.234 & 86.307
& $\mathbf{86.095 \pm 0.306}$
& 85.073 & 83.897 & 85.962
& $84.977 \pm 1.036$
& \textbf{+1.118} \\
\hline
\end{tabular}
}
\end{table}

\begin{table}[H]
\centering
\caption{
Comparison of the \texttt{3H6L1G\_steps16\_shared} two-state variant with the
\texttt{tokenizer\_only} baseline across the eight ablation datasets.
Results report classification accuracy (\%) for each seed and the mean
$\pm$ sample standard deviation across three seeds. In the Overall row, datasets are averaged equally within each seed before computing the mean and standard deviation. $\Delta$ denotes the difference
in mean accuracy between the shared-network two-state variant and the tokenizer-only baseline.
}
\label{tab:trm-vs-tokenizer}
\resizebox{\linewidth}{!}{
\begin{tabular}{lccccccccr}
\hline
& \multicolumn{4}{c}{\textbf{\texttt{3H6L1G\_steps16\_shared}}}
& \multicolumn{4}{c}{\textbf{\texttt{tokenizer\_only}}}
& \\
\cline{2-5}\cline{6-9}
\textbf{Dataset}
& \textbf{Seed 0}
& \textbf{Seed 1}
& \textbf{Seed 2}
& \textbf{Mean $\pm$ Std.}
& \textbf{Seed 0}
& \textbf{Seed 1}
& \textbf{Seed 2}
& \textbf{Mean $\pm$ Std.}
& \textbf{$\Delta$} \\
\hline

CharacterTrajectories
& 99.304 & 98.189 & 98.886
& $\mathbf{98.793 \pm 0.563}$
& 95.891 & 96.031 & 96.379
& $96.100 \pm 0.251$
& +2.693 \\

PenDigits
& 98.856 & 98.571 & 98.370
& $\mathbf{98.599 \pm 0.244}$
& 98.485 & 98.113 & 98.228
& $98.275 \pm 0.190$
& +0.324 \\

SpokenArabicDigits
& 97.954 & 98.090 & 98.363
& $\mathbf{98.136 \pm 0.208}$
& 96.362 & 96.771 & 97.317
& $96.817 \pm 0.479$
& +1.319 \\

JapaneseVowels
& 98.108 & 96.486 & 96.486
& $\mathbf{97.027 \pm 0.936}$
& 96.757 & 97.027 & 96.757
& $96.847 \pm 0.156$
& +0.180 \\

NATOPS
& 82.778 & 77.778 & 95.556
& $85.371 \pm 9.168$
& 89.444 & 92.778 & 93.333
& $\mathbf{91.852 \pm 2.103}$
& -6.481 \\

Heartbeat
& 75.122 & 70.244 & 72.195
& $72.520 \pm 2.455$
& 73.659 & 77.561 & 71.707
& $\mathbf{74.309 \pm 2.981}$
& -1.789 \\

LSST
& 66.464 & 65.815 & 67.843
& $\mathbf{66.707 \pm 1.036}$
& 66.869 & 66.504 & 64.882
& $66.085 \pm 1.058$
& +0.622 \\

DuckDuckGeese
& 62.000 & 66.000 & 60.000
& $\mathbf{62.667 \pm 3.055}$
& 62.000 & 56.000 & 56.000
& $58.000 \pm 3.464$
& +4.667 \\

\hline
\textbf{Overall}
& 85.073 & 83.897 & 85.962
& $\mathbf{84.977 \pm 1.036}$
& 84.933 & 85.098 & 84.325
& $84.786 \pm 0.407$
& \textbf{+0.191} \\
\hline
\end{tabular}
}
\end{table}

\newpage

\subsection{Recursive Dynamics and Inference-Time Scaling Analysis}
\label{app:inference_scaling}
This appendix documents the experimental protocol, complete measurements, and
validity diagnostics underlying Section~\ref{sec:inference_scaling}. The
interpretation of these results is given there; here we report what was
measured and how.

\subsubsection{Inference-time probe}
\label{app:inference_scaling:protocol}
Following the inference-time scaling formulations described in
Section~\ref{sec:bg:rrm_inference_scaling}, we evaluate \gls{rtsm} using stochastic
perturbations only at inference. The checkpoints are not retrained with the
randomized initialization (RI) and noise injection (NI) used by \gls{eqr}, and our RI+NI probe is not an exact \gls{ptrm}
reproduction because it additionally randomizes the initial state.

All experiments use the validation-protocol checkpoints described in
Appendix~\ref{appendix:experimental_setup} and evaluate the official UEA test
partitions. Runs completed for 28 of the 30 UEA datasets; EigenWorms (EW) and
InsectWingbeat (IW) did not complete within the allocated compute budget and
are excluded. All macro-averaged quantities weight these 28 datasets equally. Normalization differs between this probe and some historical benchmark runs. The depth--breadth comparisons therefore characterize the fixed checkpoints within the probe's preprocessing pipeline.

For each test sample we record one deterministic trajectory and $B_{\max}=512$
stochastic trajectories, each evaluated for up to $D=64$ recursive inference
steps, each comprising three recursive cycles. The deterministic trajectory uses the checkpoint's standard initial
state without perturbation. Each stochastic trajectory uses a fresh randomized
initialization with $\sigma_{\mathrm{RI}}=1.0$, followed by noise injection
between recursive steps with $\sigma_{\mathrm{NI}}=0.01$. Since the final \gls{rtsm}
is single-state, both interventions act on its recursive latent state. A
single maximum-budget run per dataset is sufficient, because the trajectory at
any smaller depth $d<D$ is a prefix of the recorded trajectory and any smaller
breadth $B<B_{\max}$ is a subset of the recorded rollouts; all $(D,B)$ combinations
are therefore evaluated offline from the same run.

For a sample $i$, rollout $b$, and recursion step $d$, let
$\hat{c}_{i,b}^{(d)}$ denote the predicted class, with true label $c_i$ and $n_{\mathrm{test}}$ test samples. Oracle pass@$B$ accuracy is
\begin{equation}
    \operatorname{pass@}B(D)
    =
    \frac{1}{n_{\mathrm{test}}}
    \sum_{i=1}^{n_{\mathrm{test}}}
    \mathbb{I}
    \left[
        \exists\, b \leq B:
        \hat{c}_{i,b}^{(D)} = c_i
    \right].
    \label{eq:eqr-pass-b}
\end{equation}
This is an \emph{oracle upper bound}: it reports whether at least one of the
first $B$ stochastic trajectories reaches the correct class, without requiring
that this trajectory be identifiable at inference time. We define
\emph{headroom} as oracle pass@$B$ accuracy minus deterministic accuracy at
the same depth.

To characterize state stabilization, we record the per-sample \gls{rms} state-update magnitude
\begin{equation}
    r_d
    =
    \sqrt{
        \frac{1}{M}
        \left\|
            y^{(d)}-y^{(d-1)}
        \right\|_F^2
    },
    \label{eq:eqr-residual}
\end{equation}
where $y^{(d)}$ is the latent state after step $d$, $M$ is its number of scalar elements, and $\|\cdot\|_F$ is the Frobenius norm. This is a state-stabilization
diagnostic rather than a task metric.

Three label-free selection rules are evaluated at the maximum inference
budget. \emph{Best-Q} selects the trajectory with the largest halting-head
score, \emph{majority} returns the modal class across trajectories, and
\emph{minimum residual} selects the trajectory with the smallest residual
averaged over its final three recursive steps.

\paragraph{Validity check.}
Stochastic trajectories are evaluated in expanded batches for computational
efficiency, whereas the deterministic trajectory is evaluated normally. To
exclude batch-composition artifacts, we repeat the deterministic trajectory in
the expanded rollout layout with RI and NI disabled and require identical
predictions at every recursion step. This zero-noise equivalence check passed
on all 28 datasets, with zero prediction mismatches across more than
$50{,}000$ checked sample--trajectory rows.

\subsubsection{Full depth--breadth grid}
\label{app:inference_scaling:breadth-depth}
\begin{figure}[H]
    \centering

    \begin{minipage}[t]{0.49\textwidth}
        \centering
        \includegraphics[width=\linewidth]
        {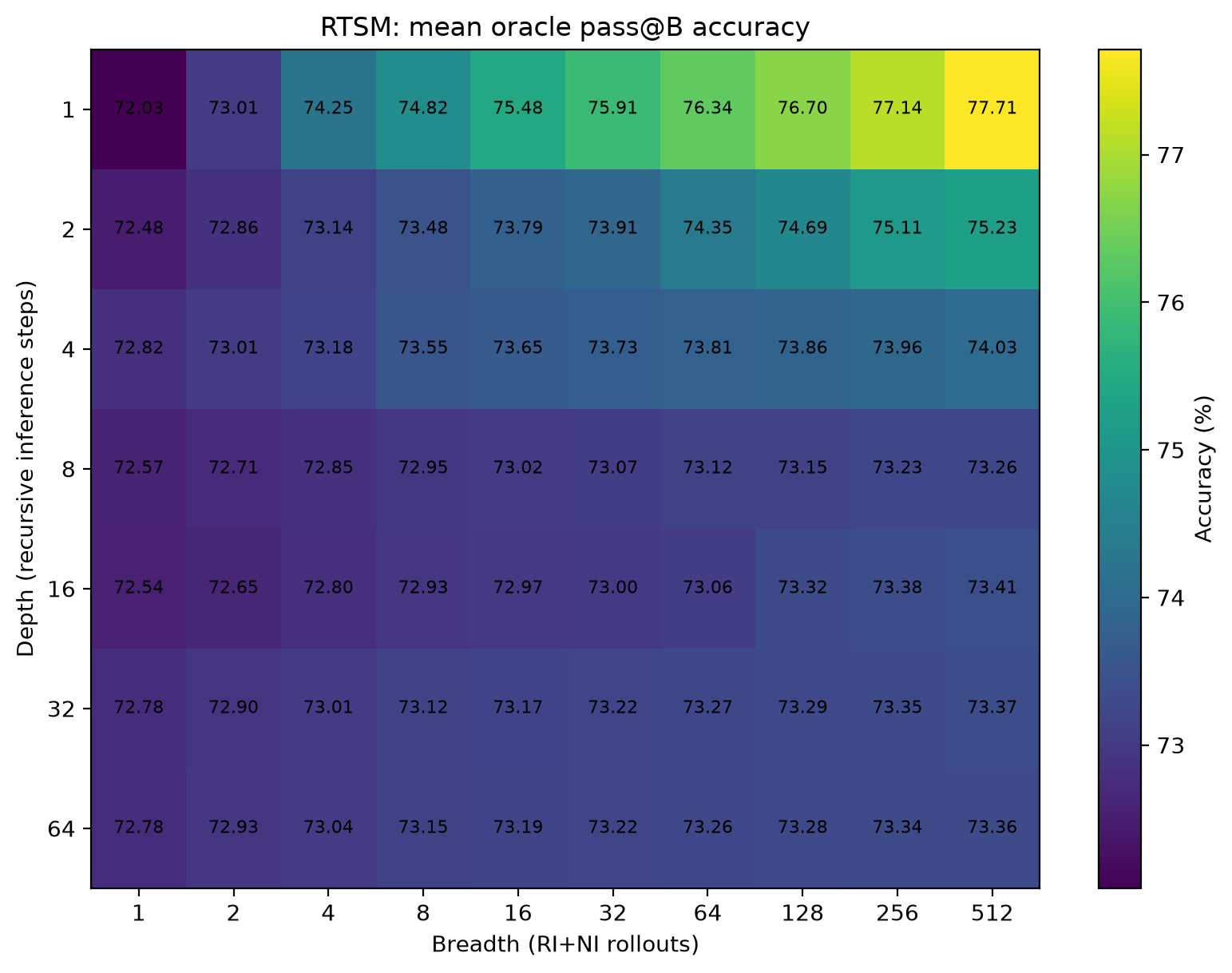}

        \vspace{1mm}
        \small (a) Oracle pass@$B$ accuracy.
    \end{minipage}
    \hfill
    \begin{minipage}[t]{0.49\textwidth}
        \centering
        \includegraphics[width=\linewidth]
        {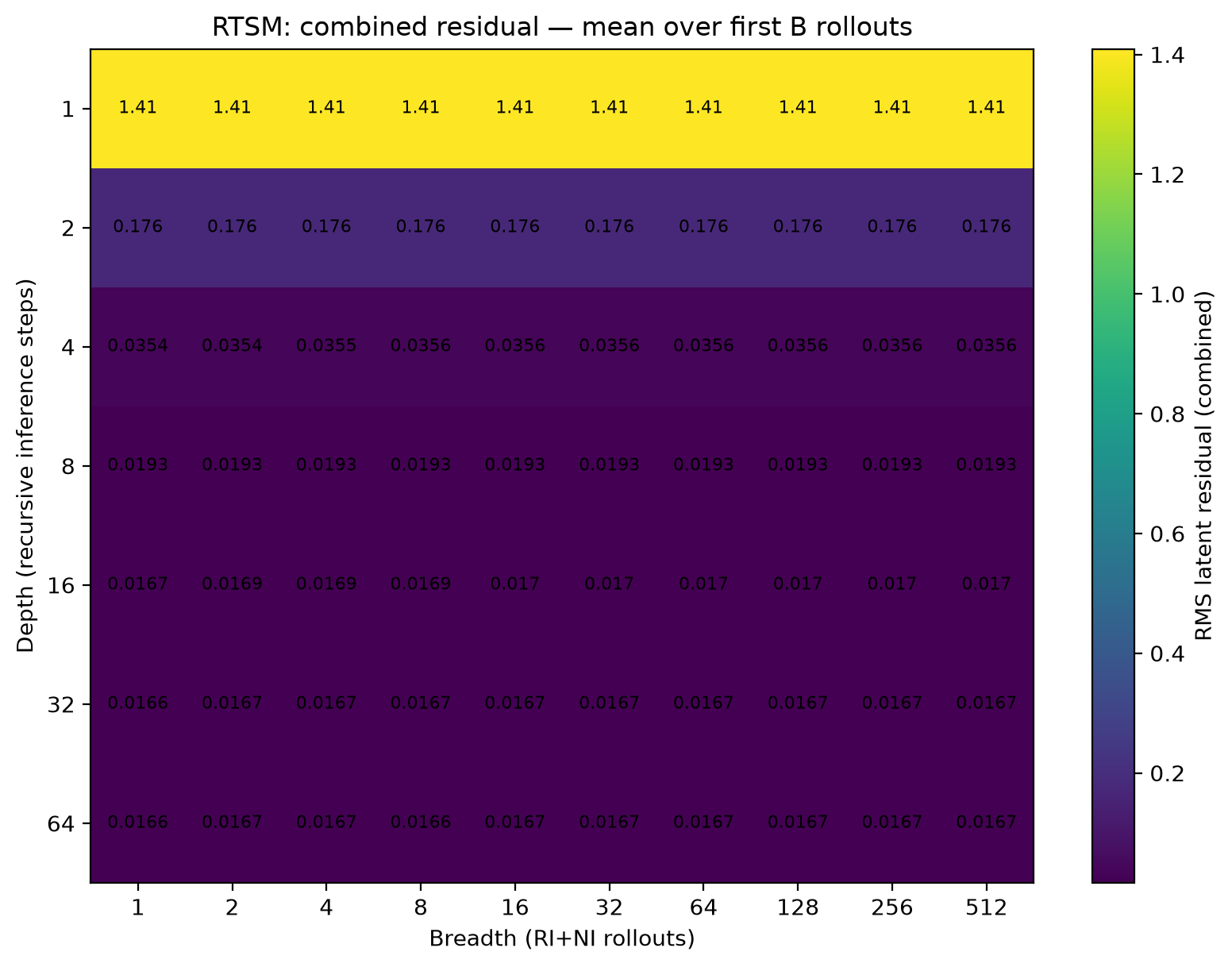}

        \vspace{1mm}
        \small (b) Mean latent-state residual.
    \end{minipage}

    \caption{
    \textbf{Complete breadth--depth grid for \gls{rtsm}.}
    Macro-averaged oracle pass@$B$ accuracy and RMS latent-state residual for
    stochastic RI+NI trajectories across 28 UEA datasets.
    }
    \label{fig:eqr-heatmaps}
\end{figure}

Figure~\ref{fig:eqr-heatmaps} and Table~\ref{tab:eqr-grid-accuracy} report the
complete $7\times10$ grid summarised in Section~\ref{sec:inference_scaling}.
The full grid supplements the selected budgets reported in the main text.
At $B=512$,
oracle accuracy falls from $77.71\%$ at $D=1$ to
$75.23\%$ at $D=2$ and $74.03\%$ at $D=4$, after which values vary within a
narrow band without a consistent trend. Across the full grid, the highest
oracle accuracy occurs at $D=1$, $B=512$ rather than at the largest joint
depth--breadth budget.

The residual grid is comparatively simple. The mean stochastic residual falls
from approximately $1.409$ at $D=1$ to $0.176$ at $D=2$, $0.0356$ at $D=4$,
and $0.0193$ at $D=8$, settling near $0.0167$ for $D\geq16$. At every fixed
depth, changing breadth from $B=1$ to $B=512$ leaves the mean residual
essentially unchanged. In this probe, depth has the larger measured effect
on the average state-update magnitude.

\begin{table}[H]
    \centering
    \scriptsize
    \setlength{\tabcolsep}{3.2pt}
    \caption{
    Macro-averaged oracle pass@$B$ accuracy (\%) across the full
    depth--breadth grid, over 28 UEA datasets.
    }
    \label{tab:eqr-grid-accuracy}
    \begin{tabular}{r|rrrrrrrrrr}
        \toprule
        $D \backslash B$
        & 1 & 2 & 4 & 8 & 16 & 32 & 64 & 128 & 256 & 512 \\
        \midrule
        1
        & 72.03 & 73.01 & 74.25 & 74.82 & 75.48
        & 75.91 & 76.34 & 76.70 & 77.14 & \textbf{77.71} \\

        2
        & 72.48 & 72.86 & 73.14 & 73.48 & 73.79
        & 73.91 & 74.35 & 74.69 & 75.11 & 75.23 \\

        4
        & 72.82 & 73.01 & 73.18 & 73.55 & 73.65
        & 73.73 & 73.81 & 73.86 & 73.96 & 74.03 \\

        8
        & 72.57 & 72.71 & 72.85 & 72.95 & 73.02
        & 73.07 & 73.12 & 73.15 & 73.23 & 73.26 \\

        16
        & 72.54 & 72.65 & 72.80 & 72.93 & 72.97
        & 73.00 & 73.06 & 73.32 & 73.38 & 73.41 \\

        32
        & 72.78 & 72.90 & 73.01 & 73.12 & 73.17
        & 73.22 & 73.27 & 73.29 & 73.35 & 73.37 \\

        64
        & 72.78 & 72.93 & 73.04 & 73.15 & 73.19
        & 73.22 & 73.26 & 73.28 & 73.34 & 73.36 \\
        \bottomrule
    \end{tabular}
\end{table}

\subsubsection{Deterministic recursive dynamics}
\label{app:inference_scaling:deterministic}
\begin{figure}[H]
    \centering

    \begin{minipage}[t]{0.32\textwidth}
        \centering
        \includegraphics[width=\linewidth]
        {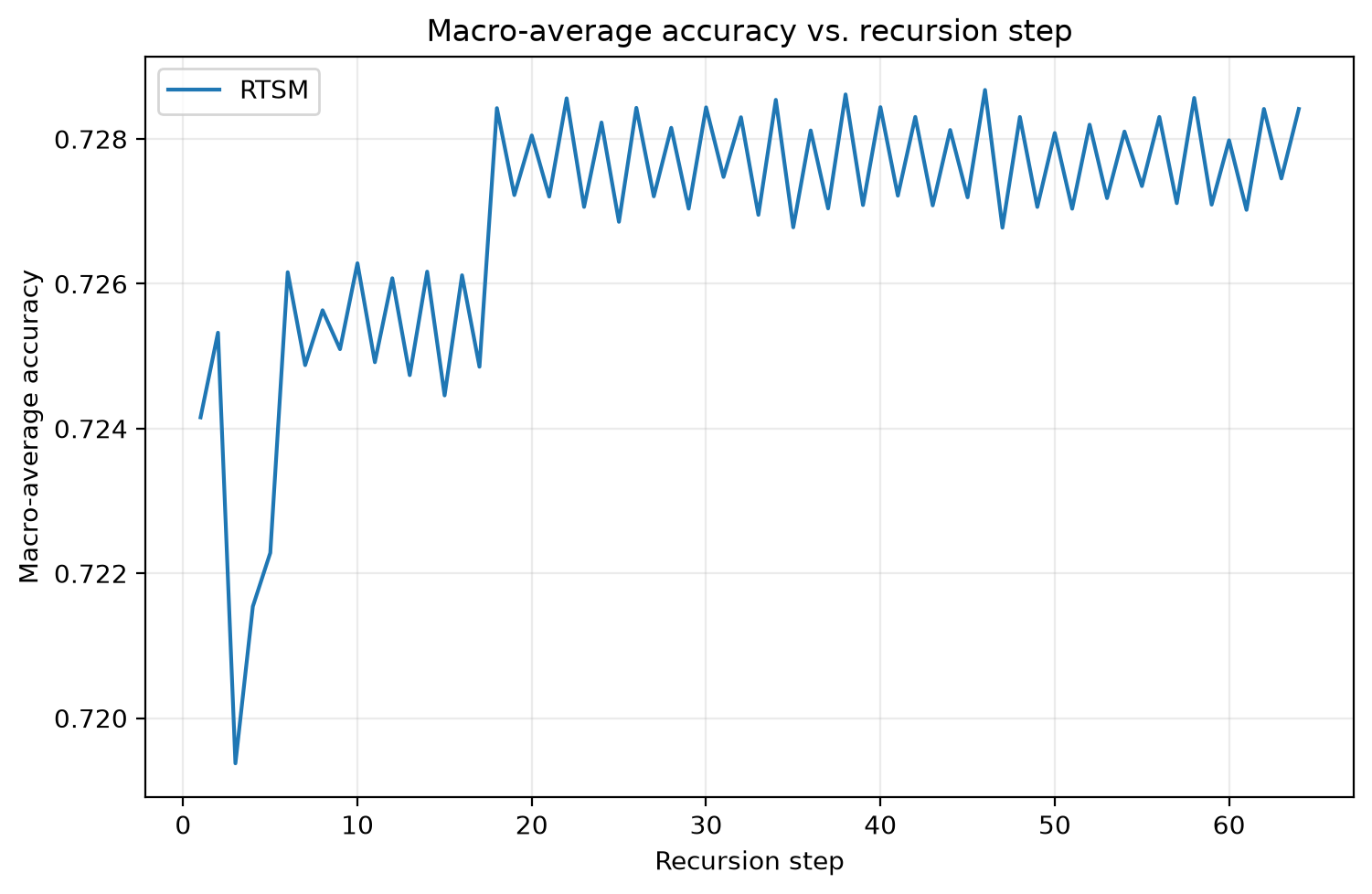}

        \vspace{1mm}
        \small (a) Accuracy.
    \end{minipage}
    \hfill
    \begin{minipage}[t]{0.32\textwidth}
        \centering
        \includegraphics[width=\linewidth]
        {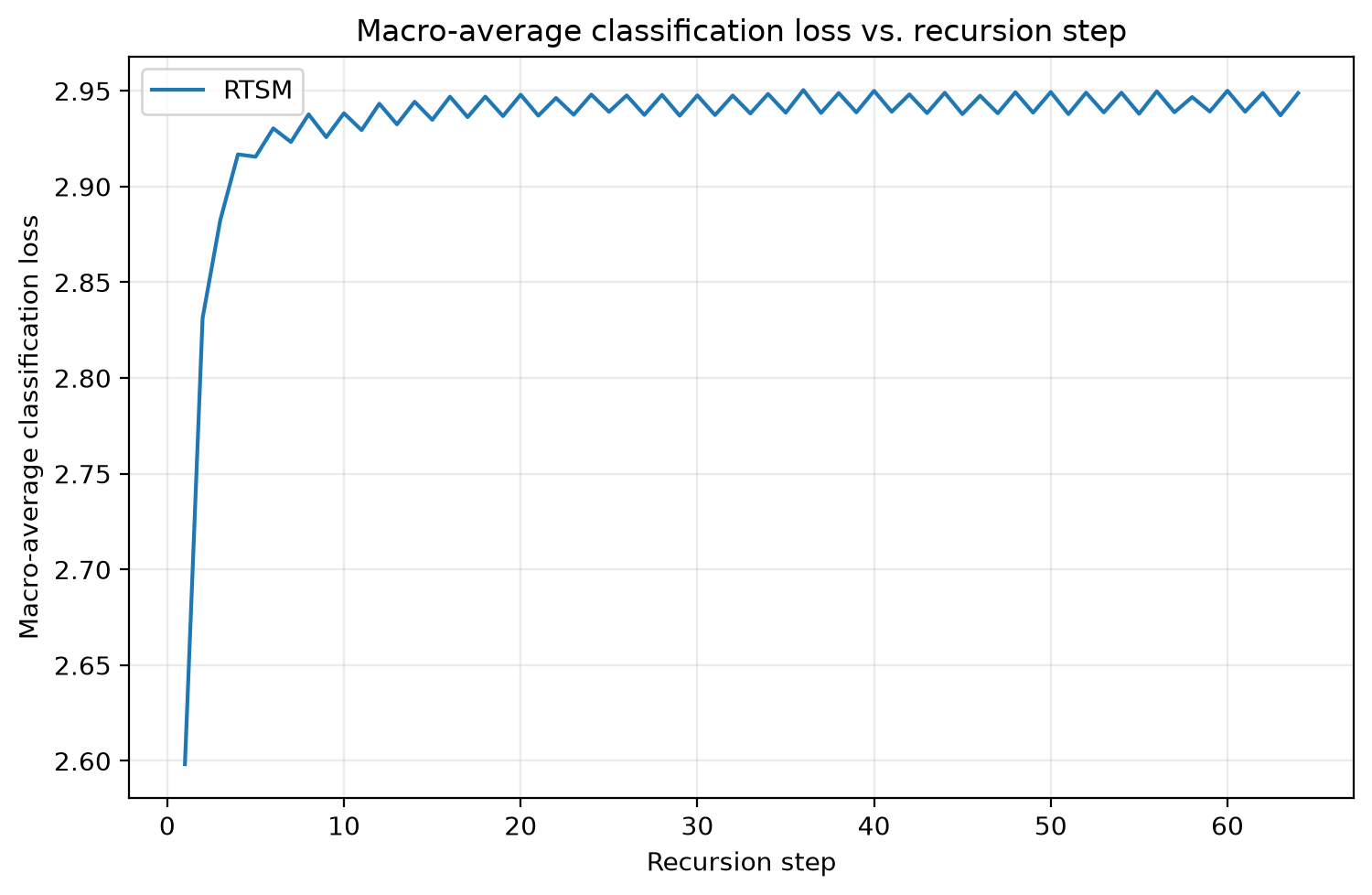}

        \vspace{1mm}
        \small (b) Cross-entropy loss.
    \end{minipage}
    \hfill
    \begin{minipage}[t]{0.32\textwidth}
        \centering
        \includegraphics[width=\linewidth]
        {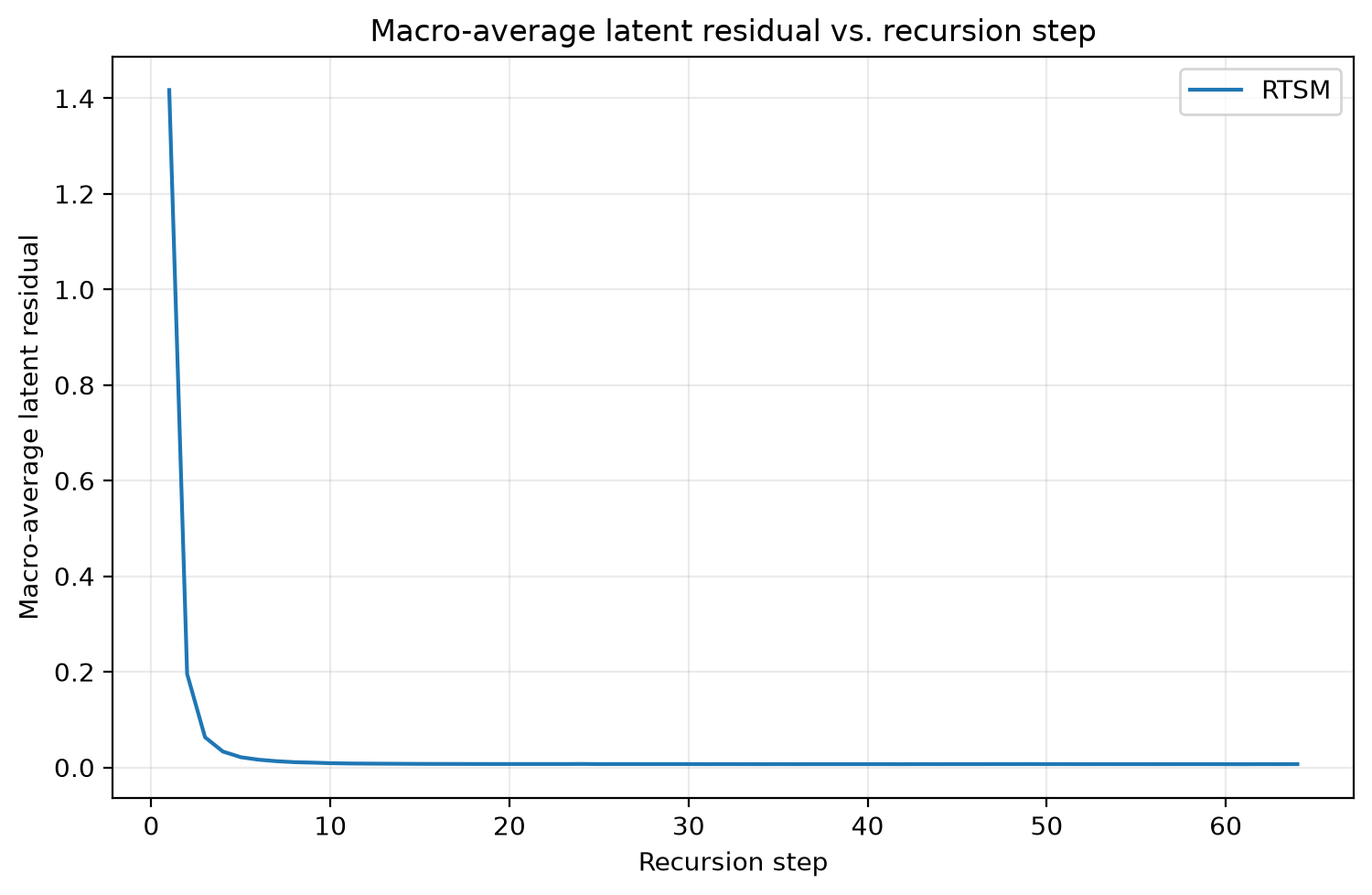}

        \vspace{1mm}
        \small (c) RMS latent residual.
    \end{minipage}

    \caption{
    \textbf{Deterministic recursive dynamics of \gls{rtsm}.}
    Macro-averaged classification accuracy, cross-entropy loss, and RMS
    latent-state residual across 64 recursion steps on 28 UEA datasets.
    }
    \label{fig:eqr-step-metrics}
\end{figure}

Figure~\ref{fig:eqr-step-metrics} reports the full 64-step traces of
classification accuracy, cross-entropy loss, and latent residual along the
deterministic trajectory; Table~\ref{tab:eqr-step-metrics} lists selected
steps. The residual decreases from $1.4172$ at the first step to $0.0330$ at
step~4, $0.0108$ at step~8, and $0.0067$ at step~64, with most of the
decrease occurring within the first few steps. Accuracy is already
$72.42\%$ after the first step and fluctuates within roughly half a percentage
point thereafter. Cross-entropy is reported as an additional diagnostic; it
increases overall across the trajectory, which provides no evidence that
further refinement improves the probabilistic predictions, but we do not
draw further conclusions from it here.

\begin{table}[H]
    \centering
    \small
    \caption{
    Deterministic trajectory metrics at selected recursion steps,
    macro-averaged over 28 datasets.
    }
    \label{tab:eqr-step-metrics}
    \begin{tabular}{rrrr}
        \toprule
        Step & Accuracy (\%) & CE loss & RMS residual \\
        \midrule
         1 & 72.42 & 2.5983 & 1.41722 \\
         2 & 72.53 & 2.8313 & 0.19541 \\
         4 & 72.15 & 2.9168 & 0.03303 \\
         8 & 72.56 & 2.9377 & 0.01080 \\
        16 & 72.61 & 2.9470 & 0.00716 \\
        32 & 72.83 & 2.9476 & 0.00684 \\
        64 & 72.84 & 2.9488 & 0.00673 \\
        \bottomrule
    \end{tabular}
\end{table}

\subsubsection{Reachable headroom and trajectory selection}
\label{app:inference_scaling:headroom}
Headroom is oracle pass@$B$ accuracy minus deterministic accuracy at the same
depth. Table~\ref{tab:eqr-headroom-macro} reports it at the maximum evaluated
budget alongside the three label-free selectors, and
Table~\ref{tab:eqr-headroom-dist} reports its distribution across datasets.
The two tables serve different purposes: the macro table places the oracle
bound and all selectors on a common footing, whereas the distribution shows
that the macro average is not representative of a typical dataset.

The Best-Q column is particularly relevant to \gls{ptrm}, which selects among
stochastic trajectories using the model's existing halting head. It therefore
tests whether the same type of Q-based selection can exploit the stochastic
headroom available under \gls{rtsm}'s learned dynamics.

\begin{table}[H]
    \centering
    \small
    \caption{
    Macro-averaged accuracy (\%) and headroom (pp) at the maximum evaluated
    budget, $D=64$ and $B=512$. Oracle pass@$B$ uses ground-truth labels and
    is an upper bound; Best-Q, majority, and minimum-residual selection are
    label-free.
    }
    \label{tab:eqr-headroom-macro}
    \begin{tabular}{lrrrrrr}
        \toprule
        & Det. & Oracle & Headroom & Best-Q & Majority & Min-res. \\
        \midrule
        \gls{rtsm}
        & 72.84 & 73.36 & +0.52 & 72.66 & 72.76 & 72.74 \\
        \bottomrule
    \end{tabular}
\end{table}

\begin{table}[H]
    \centering
    \small
    \caption{
    Distribution of per-dataset headroom (oracle pass@$B$ minus deterministic
    accuracy) at $D=64$, $B=512$, over 28 UEA datasets.
    }
    \label{tab:eqr-headroom-dist}
    \begin{tabular}{lr}
        \toprule
        Statistic & Headroom (pp) \\
        \midrule
        Mean                 & +0.52 \\
        Median               & 0.00 \\
        Q25                  & 0.00 \\
        Q75                  & +0.37 \\
        IQR                  & 0.37 \\
        Minimum              & 0.00 \\
        Maximum              & +5.40 \\
        \midrule
        Datasets $>0$        & 9 / 28 \\
        Datasets $=0$        & 19 / 28 \\
        Datasets $<0$        & 0 / 28 \\
        \bottomrule
    \end{tabular}
\end{table}

Table~\ref{tab:eqr-headroom-per-dataset} gives the per-dataset breakdown. The
positive headroom is concentrated in PhonemeSpectra and EthanolConcentration,
whose deterministic accuracies are among the lowest across the 28 evaluated
datasets; all other positive gains are below one percentage point.

\begin{table}[H]
    \centering
    \scriptsize
    \setlength{\tabcolsep}{4.0pt}
    \caption{
    Per-dataset results at $D=64$, $B=512$, sorted by headroom. All values are
    classification accuracy (\%) except headroom, reported in percentage
    points.
    }
    \label{tab:eqr-headroom-per-dataset}

    \begin{tabular}{lrrrrrr}
        \toprule
        Dataset
        & Det.
        & Oracle
        & Headroom
        & Best-Q
        & Majority
        & Min-res. \\
        \midrule

        PS   & 24.16 & 29.56 & +5.40 & 23.08 & 23.89 & 23.53 \\
        EC   & 29.66 & 34.98 & +5.32 & 25.86 & 28.52 & 28.52 \\
        FD   & 67.22 & 68.22 & +0.99 & 66.97 & 66.91 & 67.00 \\
        EP   & 94.93 & 95.65 & +0.72 & 95.65 & 94.93 & 94.93 \\
        NATO & 95.56 & 96.11 & +0.56 & 95.56 & 95.56 & 95.56 \\
        LSST & 65.37 & 65.90 & +0.53 & 65.41 & 65.45 & 65.41 \\
        PD   & 98.74 & 99.14 & +0.40 & 98.80 & 98.68 & 98.60 \\
        SAD  & 98.36 & 98.73 & +0.36 & 98.68 & 98.36 & 98.14 \\
        CT   & 99.30 & 99.51 & +0.21 & 99.03 & 99.16 & 99.16 \\

        \midrule

        AF   & 33.33 & 33.33 & 0.00 & 33.33 & 33.33 & 33.33 \\
        AWR  & 98.67 & 98.67 & 0.00 & 98.33 & 98.33 & 98.33 \\
        BM   & 100.00 & 100.00 & 0.00 & 100.00 & 100.00 & 100.00 \\
        CR   & 97.22 & 97.22 & 0.00 & 97.22 & 97.22 & 97.22 \\
        DDG  & 62.00 & 62.00 & 0.00 & 62.00 & 62.00 & 62.00 \\
        ER   & 88.52 & 88.52 & 0.00 & 88.52 & 88.52 & 88.52 \\
        FM   & 55.00 & 55.00 & 0.00 & 55.00 & 55.00 & 55.00 \\
        HB   & 68.78 & 68.78 & 0.00 & 68.78 & 68.78 & 68.78 \\
        HMD  & 58.11 & 58.11 & 0.00 & 58.11 & 58.11 & 58.11 \\
        HW   & 47.41 & 47.41 & 0.00 & 47.41 & 47.41 & 47.41 \\
        JV   & 98.11 & 98.11 & 0.00 & 97.84 & 98.11 & 98.11 \\
        LIB  & 87.22 & 87.22 & 0.00 & 87.22 & 87.22 & 87.22 \\
        MI   & 54.00 & 54.00 & 0.00 & 54.00 & 54.00 & 54.00 \\
        PEMS & 87.86 & 87.86 & 0.00 & 87.86 & 87.86 & 87.86 \\
        RS   & 84.87 & 84.87 & 0.00 & 84.87 & 84.87 & 84.87 \\
        SRS1 & 80.89 & 80.89 & 0.00 & 80.89 & 80.89 & 80.89 \\
        SRS2 & 51.67 & 51.67 & 0.00 & 51.67 & 51.67 & 51.67 \\
        SWJ  & 26.67 & 26.67 & 0.00 & 26.67 & 26.67 & 26.67 \\
        UW   & 85.94 & 85.94 & 0.00 & 85.62 & 85.94 & 85.94 \\

        \midrule
        \textbf{Mean}
        & \textbf{72.84}
        & \textbf{73.36}
        & \textbf{+0.52}
        & \textbf{72.66}
        & \textbf{72.76}
        & \textbf{72.74} \\
        \bottomrule
    \end{tabular}
\end{table}

\subsubsection{Scope and relation to prior inference-time scaling}
\label{app:inference_scaling:discussion}
\gls{eqr} applies RI and
NI during training to shape the learned dynamics, whereas the \gls{rtsm}
checkpoints evaluated here were trained without those
interventions~\citep{huang2026eqr}; whether such training would induce
similar scaling in time-series classification remains untested.
The setting is closer to \gls{ptrm}, which perturbs only at
inference~\citep{parviz2026ptrm}, but is not an exact reproduction: our probe
randomizes the initial state in addition to injecting noise along the
trajectory.

For these checkpoints and perturbations, greater breadth adds little oracle coverage at the largest depth, and the tested selectors do not improve accuracy. These findings concern the evaluated probe. Follow-ups include \gls{eqr}-style training and NI-only inference matching \gls{ptrm}'s intervention.

\clearpage
\section{RTSM Implementation Details}
\label{appendix:rtsm_implementation}

\subsection{Tokenizer Computations}
\label{appendix:rtsm_tokenizer}

The tokenizer combines channel projection, gated residual temporal filtering, densely connected dilated convolutions, and a final patch projection.
A learned affine map first gives
$U_0=\phi_C(\tilde{X}^{\top})\in\mathbb{R}^{T\times d}$.
For gated block $j$, the updates are
\begin{equation}
    \begin{aligned}
    (g_j,v_j)
    &=\operatorname{split}\!\left(
        \phi^{\mathrm{in}}_j(\operatorname{LN}(U_{j-1}))\right),\\
    U_j
    &=U_{j-1}+\phi^{\mathrm{out}}_j\!\left(
        \operatorname{SiLU}(g_j)\odot
        \operatorname{DWConv}_{k_g}(v_j)\right).
    \end{aligned}
    \label{eq:rtsm_gated_tokenizer}
\end{equation}
The input projection maps $d$ features to $2d$ features, which are split into the two branches; the output projection maps back to $d$ features. The depthwise convolution filters each feature along time, while the pointwise projections mix features.

Let $V_0=U_J^\top\in\mathbb{R}^{d\times T}$. Dilated branch $i$ computes
\begin{equation}
    V_i=\operatorname{GELU}\!\left(
        \operatorname{Conv}_{k_{\mathrm{dil}},\delta_i}
        ([V_0;\ldots;V_{i-1}])\right),
    \qquad i=1,\ldots,I,
    \label{eq:rtsm_dilated_tokenizer}
\end{equation}
where concatenation is along the feature dimension and the full convolution returns $d$ features. Each branch therefore combines the preceding feature maps with temporal context at dilation $\delta_i$. These convolutions preserve temporal length. The map $V=V_I$ is passed to the unpadded strided projection in Eq.~\ref{eq:rtsm_patch_tokenizer}. The reference tokenizer uses $J=2$ gated blocks with kernel size $k_g=5$ and $I=2$ densely connected dilated branches with kernel size $k_{\mathrm{dil}}=3$ and dilations $\delta_1=1$ and $\delta_2=2$.

\subsection{Training Settings}
\label{appendix:rtsm_training}

The benchmark configuration uses AdamW with learning rate $10^{-4}$, $(\beta_1,\beta_2)=(0.9,0.95)$, and weight decay $1$. The learning rate is constant, with no warm-up, and dropout is zero. The nominal training budget is 100 epochs, with batch size selected per dataset. Protocol-specific early stopping and checkpoint selection are described in Appendix~\ref{appendix:experimental_setup}; trajectory accounting is specified in Appendix~\ref{appendix:slot_buffer}.

The classification and halting heads each use four pooling channels. The halt projection is initialized with $w_q=0$ and $b_q=-5$, so the initial halt logit is negative. An exponential moving average with decay $0.999$ tracks the trainable parameters and supplies the evaluation weights. The fixed random initial latent vector is not trained. During training, a positive halt logit can end a trajectory before the maximum $S_{\max}=16$ supervision steps. With probability $p_{\mathrm{exp}}=0.1$, a minimum depth is drawn uniformly from $\{2,\ldots,S_{\max}\}$ and early halting is suppressed until that depth is reached. This exploration rule affects training only; benchmark inference always uses 16 steps.

\clearpage
\subsection{Component Comparison}
\label{appendix:rtsm_comparison}

Table~\ref{tab:rrm-components} records the relationship between RTSM and its recursive-model predecessors. The main architectural changes concern the input representation, state formulation, and sequence-level readout.

\begin{table}[H]
\centering
\caption{Component-level comparison of HRM, TRM, and RTSM using the framework
of Section~\ref{sec:bg:rrm_general}. The comparison follows the released \gls{hrm} and \gls{trm} implementations. HRM provides architectural context;
the HRM-style ablation in Section~\ref{sec:recursive_ablations} changes
state structure and network sharing within the TRM training recipe.
$H/L/G$ denotes cycles, inner auxiliary-state updates, and
gradient-tracked cycles.}
\label{tab:rrm-components}
\footnotesize
\setlength{\tabcolsep}{4pt}
\renewcommand{\arraystretch}{1.05}
\begin{tabular}{@{}
>{\raggedright\arraybackslash}p{0.190\textwidth}
>{\raggedright\arraybackslash}p{0.250\textwidth}
>{\raggedright\arraybackslash}p{0.250\textwidth}
>{\raggedright\arraybackslash}p{0.250\textwidth}
@{}}
\toprule
\textbf{Component} & \textbf{HRM}~\citep{wang2025hrm} & \textbf{TRM}~\citep{jolicoeur2025trm} & \textbf{RTSM (ours)} \\
\midrule
State initialization
  & frozen random vectors
  & frozen random vectors
  & frozen random vector ($y$ only) \\
Update networks
  & separate $f_L$, $f_H$
  & one shared network $f$
  & one shared network $R_\theta$ \\
Gradient schedule
  & one-step gradient: final $f_L$ and $f_H$ updates only
  & full backpropagation through the final cycle ($G=1$)
  & full backpropagation through the final cycle ($G=1$) \\
Deep supervision
  & up to $S_{\max}=16$ steps, states detached between steps
  & up to $S_{\max}=16$ steps, states detached between steps
  & up to $S_{\max}=16$ steps, states detached between steps \\
Halting, training
  & Q-learning halt/continue head with exploration
  & correctness-based halt score $q$ with BCE and exploration
  & same as TRM, using sequence-level correctness  \\
Halting, evaluation
  & fixed depth $S_{\mathrm{eval}}$
  & fixed depth $S_{\mathrm{eval}}$
  & fixed depth $S_{\mathrm{eval}}$ \\
Loss
  & Stablemax cross-entropy + Q-learning loss
  & Stablemax cross-entropy + $\lambda_q$ halting BCE
  & Stablemax cross-entropy + $\lambda_q$ halting BCE  \\
Parameter EMA
  & no
  & yes
  & yes \\
\midrule
Latent states
  & $z_H$, $z_L$
  & $y$, $z$
  & $y$ \\
Reference cycles
  & $H/L=2/2$
  & $H/L=3/6$
  & $H/L=3/0$ \\
Input injection
  & $X$ enters the low-level update; $z_H\leftarrow f_H(z_L+z_H)$
  & $X$ enters the $z$-update; $y\leftarrow f(y+z)$
  & $y\leftarrow R_\theta(y+X)$ in every cycle \\
Encoder $f_I$
  & discrete token + puzzle embeddings; RoPE
  & discrete token + puzzle embeddings; RoPE (attention variant)
  & convolutional time-series tokenizer, $N$ patch tokens; no positional encoding \\
Output head $f_O$
  & per-token linear head
  & per-token linear head
  & pooled sequence-level class and halt heads (Eq.~\ref{eq:rtsm_heads}) \\
Training loop
  & persistent ACT carry with halted-sample replacement
  & persistent ACT carry with halted-sample replacement
  & per-sample trajectory accounting (Section~\ref{sec:rtsm:training}) \\
\bottomrule
\end{tabular}
\end{table}

\end{document}